\documentclass[journal]{IEEEtran}

\usepackage{subfig}
\usepackage{caption}
\usepackage{float}
\usepackage{graphicx}
\usepackage{amsfonts} 
\usepackage{amssymb}
\usepackage{bm}
\usepackage{amsmath}
\usepackage{color}
\usepackage{multirow}
\usepackage[colorlinks,linkcolor=red]{hyperref}

\ifCLASSINFOpdf
\else
\fi

\begin{document}
\title{Video Person Re-identification using Attribute-enhanced Features}

\author{Tianrui~Chai,
    	Zhiyuan~Chen,
        Annan~Li$^*$,~\IEEEmembership{Member,~IEEE,}
        Jiaxin~Chen,
        Xinyu~Mei,
        and~Yunhong~Wang,~\IEEEmembership{Fellow,~IEEE}
\thanks{T. Chai, A. Li, J. Chen, X. Mei and Y. Wang are with the School of Computer Science and Engineering
, Beihang University, Beijing, China, 100191, e-mail: \{trchai,liannan,jiaxinchen,xymei,yhwang\}@buaa.edu.cn.}
\thanks{Z. Chen is with the Alibaba group, e-mail: dechen@buaa.edu.cn. This work was done when Chen was a master student at Beihang University.}%
\thanks{A. Li is the corresponding author.}%
}

\markboth{IEEE Transactions on XXXX ,~Vol.~XX, No.~X, July~2021}%
{Chai \MakeLowercase{\textit{et al.}}: Person Re-identification Attribute}

\maketitle

\begin{abstract}
Video-based person re-identification (Re-ID) which aims to associate people across non-overlapping cameras using surveillance video is a challenging task. 
Pedestrian attribute, such as gender, age and clothing characteristics contains rich and supplementary information but is less explored in video person Re-ID. 
In this work, we propose a novel network architecture named Attribute Salience Assisted Network (ASA-Net) for attribute-assisted video person Re-ID, which achieved considerable improvement to existing works by two methods.
First, to learn a better separation of the target from background, we propose to learn the visual attention from middle-level attribute instead of high-level identities.
The proposed Attribute Salient Region Enhance (ASRE) module can attend more accurately on the body of pedestrian.
Second, we found that many identity-irrelevant but object or subject-relevant factors like the view angle and movement of the target pedestrian can greatly influence the two dimensional appearance of a pedestrian.
This problem can be mitigated by investigating both identity-relevant and identity-irrelevant attributes via a novel triplet loss which is referred as the Pose~\&~Motion-Invariant (PMI) triplet loss.
%
%
%
%
%
Extensive experiments on MARS and DukeMTMC-VideoReID datasets show that our method outperforms the state-of-the-art methods. 
Also, the visualizations of learning results well-explain the mechanism of how the improvement is achieved.  
\end{abstract}


\IEEEpeerreviewmaketitle

\section{Introduction}
\IEEEPARstart{P}{erson} re-identification (Re-ID) is a specific person retrieval problem which searches a particular person in a query image or video across non-overlapping cameras. It has a wide range of applications, such as person tracking, criminal searching and activity analysis and has attracted more and more attention.

\begin{figure}[t]
    \centering
    \includegraphics[scale=0.33]{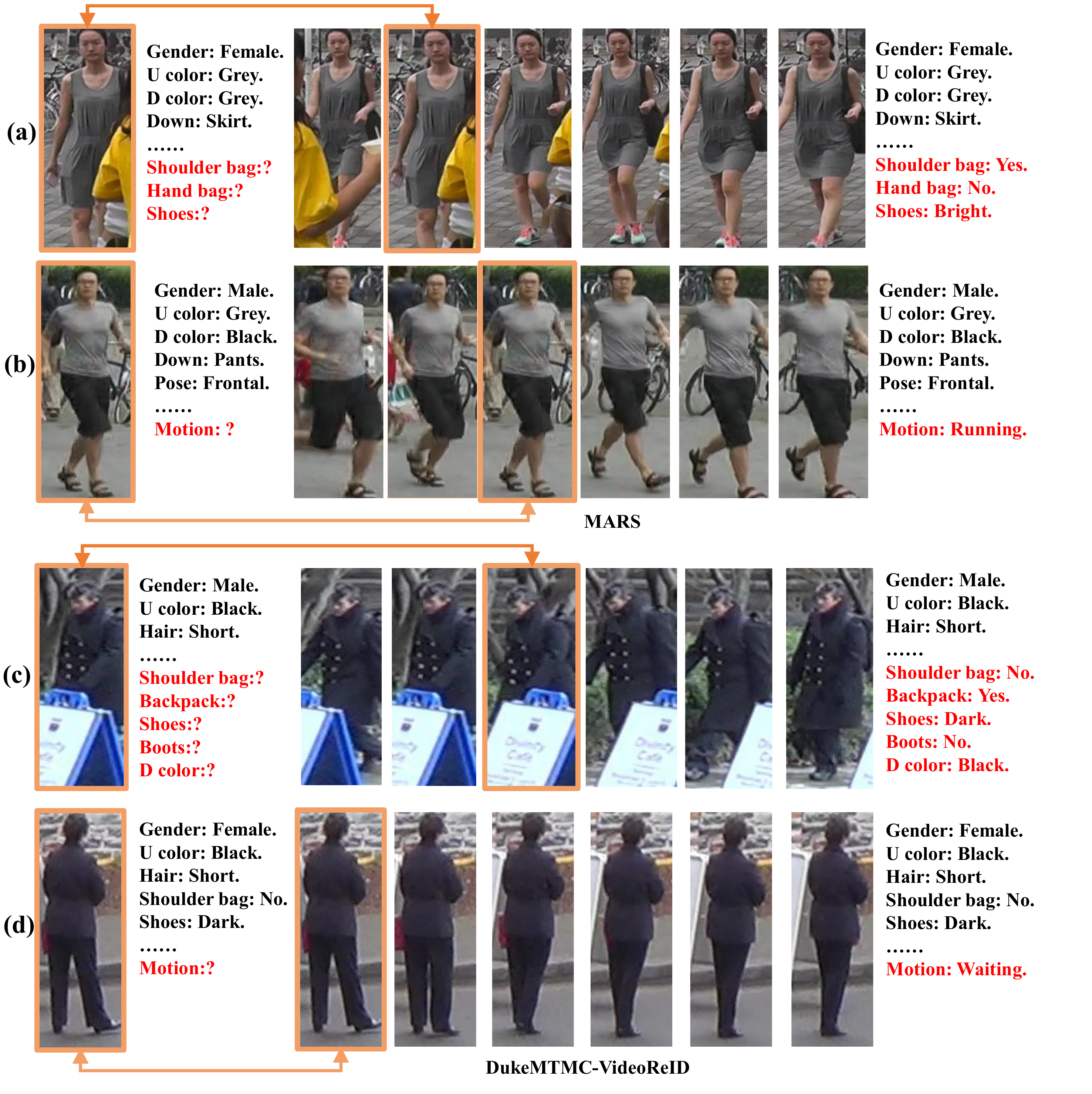}
    \setlength{\abovecaptionskip}{0pt}
    \setlength{\belowcaptionskip}{0pt}
    \caption{Exemplar sequences and corresponding attribute annotations. As shown in the left image of (a) and (c), it's hard to identify all the identity-relevant attribute from a single image. However, these attributes can be clearly observed from other frames. Despite the identity-relevant attributes, we find identity-irrelevant attributes are also helpful. Similarly from the left single image of (b), it's hard to say whether the subject is walking or running. But with the help of adjacent frames we can infer that he is running. Also, the waiting behavior of the woman in sequence (d) is ambiguous when only one image is available.}
    \label{fig:Attribute_1}
    \vspace{-0.6cm} 
\end{figure}

In the past few years, many methods have been proposed for image-based person Re-ID~\cite{bai2017scalable,hirzer2011person,kviatkovsky2012color,zheng2017person,hermans2017defense,chen2017beyond} and video-based person Re-ID~\cite{gu2020appearance,chen2020temporal,yang2020spatial,wang2021robust}. Remarkable results have been achieved with them. These methods can be divided into two categories: i.e. feature or representation learning~\cite{bai2017scalable,hirzer2011person,kviatkovsky2012color,gu2020appearance,chen2020temporal,yang2020spatial} and metric learning~\cite{zheng2017person,hermans2017defense,chen2017beyond} respectively. The former focuses on learning discriminative appearance feature of pedestrian, while the latter focuses on deriving appropriate distance metric for matching.

\begin{figure*}[t]
    \centering
    \includegraphics[width=6.3in]{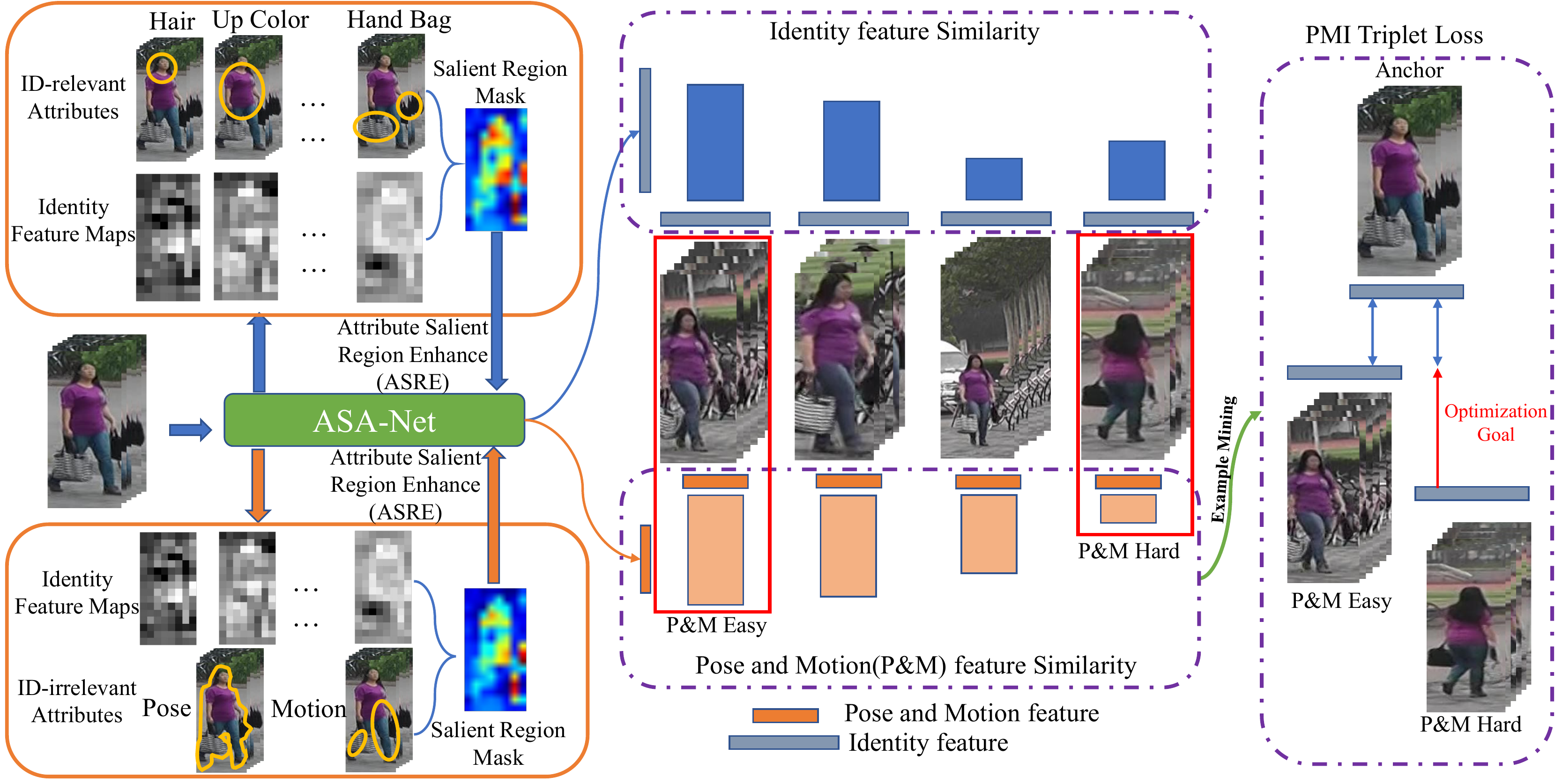}
    \setlength{\abovecaptionskip}{1pt}
    \setlength{\belowcaptionskip}{1pt}
    \caption{Overview of our methods. When training the ASA-Net, both ID-relevant and ID-irrelevant attributes are used to enhance features. Pose \& motion feature as well as the identity feature are calculated with ASA-Net. And Pose \& motion feature is used for mining robust identity feature based on the Pose \& Motion-Invariant (PMI) triplet loss.}
    \label{fig:overview}
    \vspace{-0.4cm} 
\end{figure*}

No matter the aforementioned feature learning or the metric learning, they are all supervised by the identity which is at a high level of visual semantics. Since prior arts found that middle-level attributes such as gender, age and clothing characteristics can provide additional information, they have been incorporated in person Re-ID. Existing approaches for attribute-enhanced or assisted person Re-ID are mainly based on static image. Some works~\cite{su2017multi,wang2018transferable,shi2020person} considered attributes as discriminative features while the others~\cite{lin2019improving,ling2019improving,han2018attribute,zhang2020person,tay2019aanet} take attributes as a supervision to help co-training. 

However, though pedestrian attributes show good assistance in image-based person Re-ID, we argue that attributes haven't fully play a role due to the limitations of static image. As shown in (a) and (c) of Figure~\ref{fig:Attribute_1}, attributes such as \emph{shoulder bag}, \emph{shoes} and \emph{down color} cannot be observed from a single frame due to partial occlusions. But these attributes can be clearly identified from other frames of the sequence. A video or an image sequence can provide more complete pedestrian attribute information. Consequently, exploring attribute cues in video is a better way of tackling attribute-assisted person Re-ID. Some efforts have been made to improving video person Re-ID by attributes. Song et al.~\cite{song2019two} and Zhao et al.~\cite{zhao2019attribute} first introduced attributes into video-based person Re-ID. Li et al.~\cite{li2020appearance} designed an encoder-decoder structure to capture the motion information to enhance the Re-ID model. However, such pioneer works have two obvious limitations.

First, although video or image sequence can provide redundant information to person Re-ID, there are still some problems to be solved. Specifically, besides outer factors like background clutter or temporal occlusions, the viewpoint/pose difference, movement variation or other object-relevant but identity-irrelevant (ID-irrelevant) factors can also result in a dramatic change in two dimensional appearance. Existing approaches for video-based person Re-ID seldom include any mechanism of explicitly identifying such factor. Consequently, some ID-irrelevant pattern will be inevitably treated as part of the subject-specific feature, and results in errors of identification.

Second, we observe that one of the fundamental challenge of pedestrian recognition is the template a fact that the template is usually a simple rectangle but the shape or silhouette of a person can be various. Therefore, separating background element from the target pedestrian is always an important issue. In video-based settings, the bounding box of target pedestrian is obtained by automatic detector and tracker, by which some errors are inevitable. The inaccurate box makes the target-background separation issue worse. To mitigate this problem, explicit segmentation~\cite{subramaniam2019co,choi2021arm} and visual attention~\cite{li2018diversity,fu2019sta,liu2019spatially,zhu2020asta,gong2021lag} have been adopted. The former requires a bulk of pixel-level annotation, which is difficult to obtain, and the latter is usually supervised by the identity labels. We argue that although identity labels can tell who is the people, the clue they provide about the spatial location of the people is coarse-grained. In contrast, attributes can provide various fine-grained information about the figure, motion/action and the status of carrying article of the target people. Therefore, learning from attributes is a better way of tackling the target-background separation issue.

Based on the above observations, we propose a comprehensive architecture or attribute-assisted video person re-identification. It is referred as Attribute Salience Assisted Network (ASA-Net). Its overview is shown in Figure~\ref{fig:overview}. ASA-Net combines both attribute and identity recognition and it is a five-branch structure: two attribute branches for learning attribute feature, one base branch for extracting original identity features and two branches for learning the attribute-enhanced identity feature. What makes ASA-Net different from existing method lies in two aspects: First, to address the target-background separation issue in an attribute-driven manner, we design the Attribute Salient Region Enhance (ASRE) module. By introducing this module the clutteredness caused by the background elements can be considerably reduced. Second, to deal with the ID-irrelevant but subject-relevant factors, we propose the Pose~\&~Motion-Invariant triplet loss (PMI triplet Loss). The PMI triplet Loss aims to mine the hardest samples through the distance of pose and motion states, and to reduce the intra-class distance caused by pose and motion state differences. Using the attribute annotation provided by Chen et al.~\cite{chen2019temporal} the proposed ASA-Net achieved promising performance on MARS~\cite{zheng2016mars} and DukeMTMC-VideoReID~\cite{ristani2016performance} datasets, which well demonstrate the effectiveness of our method.

In summary, our contributions are three-fold:

\begin{enumerate}
  \item  We design a five-branch network named ASA-Net which integrates video-based attribute recognition and person Re-ID. The novel Attribute Salient Region Enhance (ASRE) module can effectively focus on the attribute salient region and enhance the feature.  
  \item We propose pose \& motion-invariant triplet loss (PMI Triplet Loss) which mines the hardest samples with the midlle-level ID-irrelevant attributes to reduce the difference introduced by the change of pose and motion.
  \item We evaluate proposed ASA-Net and PMI triplet loss on MARS~\cite{zheng2016mars} and DukeMTMC-VideoReID~\cite{ristani2016performance}, on which the proposed method outperforms the state-of-the-art. The effectiveness of our method is validated by both experiments and the visualization of learning results.
\end{enumerate}

The rest of the paper is organized as follows: we overview works closely related to our work in Section~\ref{sec:related}. We elaborate the details of the proposed method in Section~\ref{sec:method}, and provide experimental results and analysis of our method by comparing with the state-of-the-art approaches in Section~\ref{sec:exp}. Finally we conclude in Section~\ref{sec:conclude}. 

\section{Related work}
\label{sec:related}

In this section, we provide a brief review of relevant existing Re-ID methods. They can be classified into three categories: image-based person Re-ID, video-based person Re-ID, and attribute-assisted person Re-ID in both image-based and video-based settings.

\subsection{Image-based Person Re-ID}

Person Re-ID is a challenging task which has been studied for years. Existing works on image-based person Re-ID can be divided into two categories: the first one focus on learning a discriminative ID-level representation for the input pedestrian image~\cite{bai2017scalable,hirzer2011person,kviatkovsky2012color,liu2012person,zhao2017deeply,gong2021lag} while the second one focus on learning robust distance metrics~\cite{xiong2014person,liao2015efficient,yan2021beyond,liu2019person}. In the early years, methods focused on the design and extraction of hand-crafted features~\cite{li2013locally,matsukawa2016hierarchical}. With the great success of deep learning~\cite{he2016deep}, convolutional neural networks have been adopted in many approaches~\cite{ahmed2015improved,varior2016gated,li2014deepreid,lin2017consistent} and have achieved great improvements. Although achieved good results, image-based settings actually avoid the negative influence of inaccurate pedestrian detection and tracking. To meet the requirements from real-world application, video-based person Re-ID which is based on automatic detection and tracking is becoming the mainstream.

\subsection{Video-based Person Re-ID}

Compared with image, temporal relation between frames in a pedestrian sequence or tracklet can make up for important identity information lost by single image due to occlusion, view change and other factors. The same as image-based person Re-ID, traditional approaches for video-based person re-id also focus on finding efficient hand-crafted descriptors~\cite{wang2014person,tao2013person}. In the past years, deep learning based approaches~\cite{mclaughlin2016recurrent,zhou2017see,xu2017jointly,liu2019spatial} for video-based person Re-ID have been proposed and show good performance. Recently, some work focus on the mining of spatial-temporal relationships. Yang et al.~\cite{yang2020spatial} and Yan et al.~\cite{yan2020learning} construct nodes according to the body part and the spatial-temporal features are aggregated by using graph convolution network. Gu et al.~\cite{gu2020appearance} put his hand to the problem of spatial misalignment of video clips and proposed the AP3D net to align the adjacent in pixel level. Zhang et al.~\cite{zhang2020multi} proposed the MGRA net to reduce redundancy within a clip.

\begin{figure*}[t]
    \centering
    \includegraphics[width=6.2in]{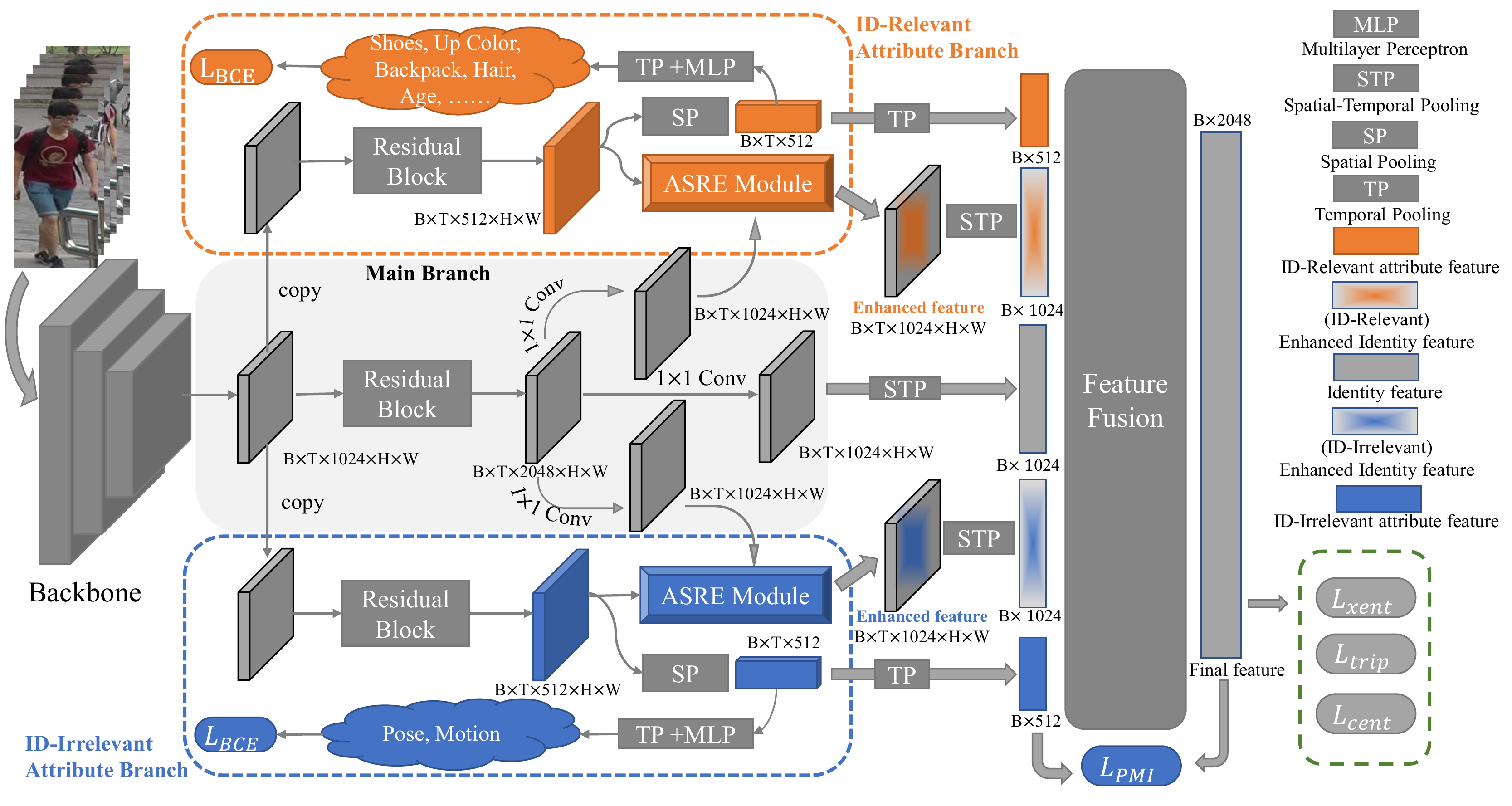}
    \setlength{\abovecaptionskip}{0pt}
    \setlength{\belowcaptionskip}{0pt}
    \caption{Overall framework of our proposed ASA-Net. The input sequence has $T$ frames and we use a CNN backbone to extract the feature map of each frame. Then we send the feature map into three branches. The middle row is the identity branch which extracts original identity features. The ID-relevant attribute branch extracts the features for recognizing attributes like gender, clothes, hair and etc. The identity features is enhanced by attribute feature using the Attribute-Salient-Region-enhance (ASRE) module. The framework of ID-irrelevant attribute branch is similar to the ID-relevant branch. Finally we fuse the obtained five features to get the final feature. At the same time, we also construct the Pose\&Motion-invariant (PMI) loss to eliminate the influence caused by pose and motion according to the ID-irrelevant attribute features. }
    \label{fig:network}
    \vspace{-0.3cm} 
\end{figure*}

\subsection{Attribute-assisted Person Re-ID}

Pedestrian attribute recognition~\cite{deng2014pedestrian,liu2017hydraplus,zhao2018grouping,gao2019pedestrian} has been a hot topic because of its practicability and great demand from video surveillance. Due to the natural similarity between pedestrian attribute learning and person Re-ID, it is natural to take them into consideration at the same time. Compared with the sole person Re-ID, attribute can provide a higher-level semantic discriminant information such as gender, ethnic and age. Some early methods use the attributes to increase the discrimination of identity feature~\cite{layne2012person,li2014clothing,su2017multi,wang2018transferable,li2019attribute}, while some other methods consider attribute as a supervision to help co-training~\cite{lin2019improving,ling2019improving,han2018attribute,zhang2020person,wang2019learning}. Also, attributes can be considered as a metric for identification~\cite{li2020structure}.

Song et al.~\cite{song2019two} first applied attribute information into video-based person Re-ID. Zhao et al.~\cite{zhao2019attribute} transferred the knowledge of attribute dataset into video-based person Re-ID. Li~\cite{li2020appearance} first introduced the concept of motion into the attributes and used attributes to enhance the video-based person Re-ID. Although these works show the effectiveness of attribute in video person Re-ID, the lack of attribute annotation limits the performance improvement and model development. Also, the attributes considered by these work are consistent with those in image-based person Re-ID, and the unique pedestrian ID-irrelevant attributes which have a great influence on the appearance are overlook. To this end, we propose a comprehensive study on attribute-assisted video person re-identification.
\vspace{-0.3cm}
\section{Method}
\label{sec:method}

As shown in Fig.~\ref{fig:network}, the framework for our proposed model consists of three branches. The upper branch is the ID-relevant attribute branch which extracting features of ID-relevant attributes (gender, shoes, hair, up color...) and make predictions of these attributes. The middle branch is the original branch which extracting appearance feature of the pedestrian for identity recognition. The bottom branch is the ID-irrelevant attribute branch. It is similar to the upper branch but the attributes are replaced with those unrelated to identity such as motion and pose. Every branch is not an isolated island. We use the proposed ASRE module to enhance the feature map extracted from the original branch on the spatial level according to the feature map extracted from the attribute branch. Five branches of features will be obtained with ASA-Net. Especially, the feature extracted from ID-irrelevant branch are used to calculate the proposed PMI triplet loss. 

In this section, we will first introduce the overall framework of our ASA-Net. The next we will introduce the ASRE module, feature fusion strategy, the PMI  triplet loss and model optimization in detail respectively.

\begin{figure*}[t]
    \centering
    \includegraphics[width=5.5in]{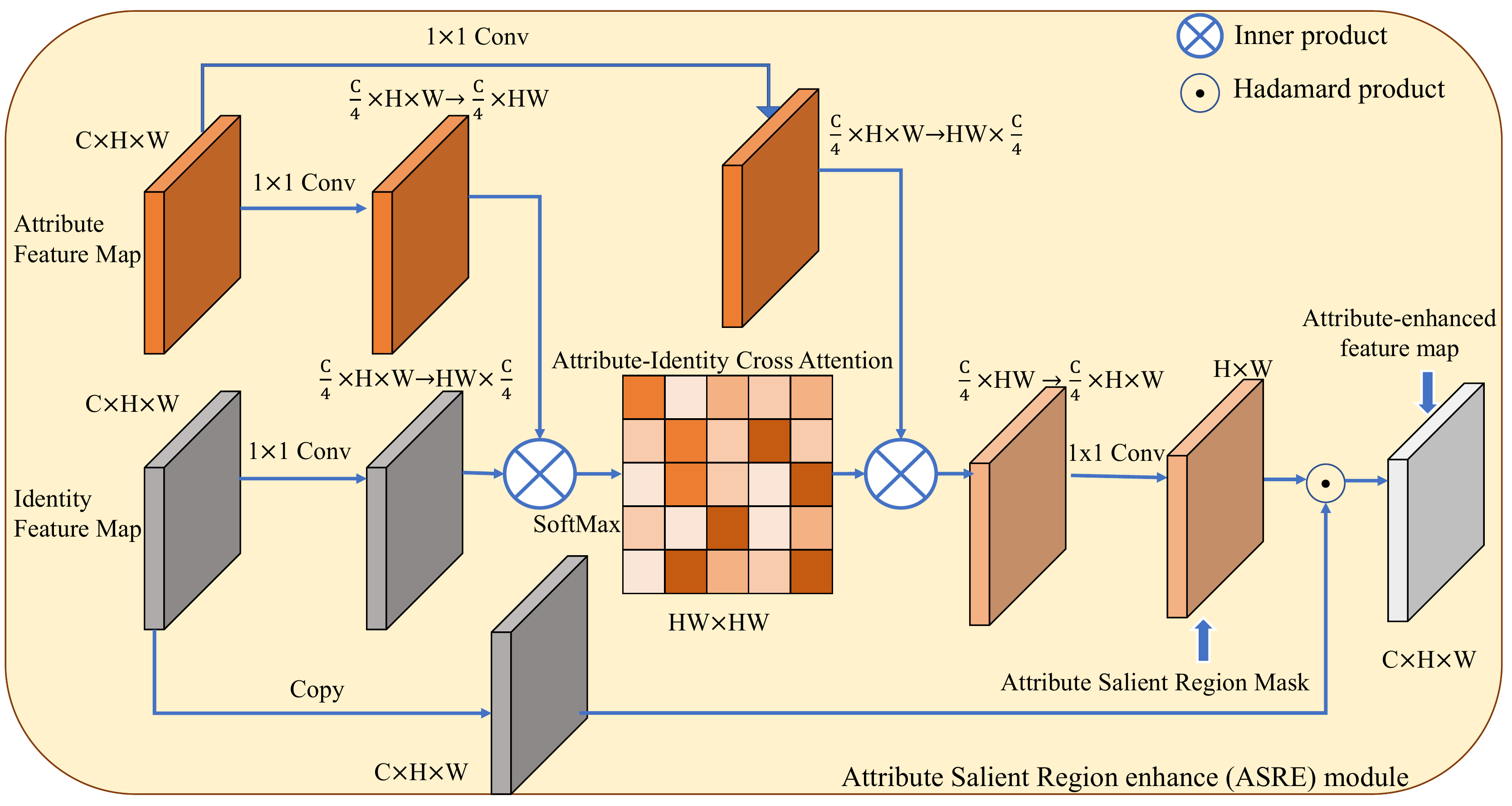}
    \setlength{\abovecaptionskip}{0pt}
    \setlength{\belowcaptionskip}{0pt}
    \caption{Illustration of the Attribute Salient Region Enhance (ASRE) module. The response degree of each pixel in the identity feature map to each pixel in the attribute feature map can be calculated by cross attention mechanism. Then the obtained Attribute-Identity attention map multiplies with the projected attribute feature map to get the response value of the attribute on each pixel of the identity feature map. Finally, the final attribute salient region mask is calculated from response value map.}
    \label{fig:ASRE}
    \vspace{-0.55cm} 
\end{figure*}

\subsection{Overall Framework}

Given a pedestrian video clip or tracklet, we denote it as $\bm{I}=\{\bm{I}_{1},\bm{I}_{2},\bm{I}_{3},...\bm{I}_{T}\}$ where $T$ is the number of frames sampled from the clip. The feature map extracted by the backbone network is $\bm{X}_{origin}=[\bm{X}_{1},\bm{X}_{2},...\bm{X}_{t}]$ where $\bm{X}_{i} \in \mathbb{R}^{C \times H \times W}$ and $\bm{X}_{origin} \in \mathbb{R}^{T \times C \times H \times W}$. We copy the feature map $\bm{X}_{origin}$ twice and send them into attribute branches. After their own residual module~\cite{he2016deep} of the three branches, we can get three basic feature maps $\bm{X}_{re\_attr}, \bm{X}_{ID}, \bm{X}_{ir\_attr}$. The process can be expressed as:
\begin{equation}
\begin{aligned}
   \bm{X}_{origin} &= F_{backbone}(\bm{I}), 
   \\
   \bm{X}_{re\_attr} &= F_{residual\_re}(\bm{X}_{origin})
   \\
   \bm{X}_{ID} &= F_{residual\_ID}(\bm{X}_{origin})
   \\
   \bm{X}_{ir\_attr} &= F_{residual\_ir}(\bm{X}_{origin}),
\end{aligned}
\end{equation}
where $\bm{X}_{re\_attr}, \bm{X}_{ir\_attr}\in\mathbb{R}^{T \times \frac{C}{2} \times H \times W}$ and $\bm{X}_{ID}\in\mathbb{R}^{T \times 2C \times H \times W}$. $F_{residual\_re}$, $F_{residual\_ID}$ and $F_{residual\_ir}$ are residual module~\cite{he2016deep} of the corresponding dimension.

In the ID-relevant attribute branch, $\bm{X}_{re\_attr}$ are used to calculated the final ID-relevant attribute feature $\bm{f}_{re\_attr}$ which is used for ID-relevant attribute prediction and the final identity feature calculation. At the same time, the enhanced feature map $\bm{X}_{re\_ID}$ is calculated with the ASRE module according to the ID-relevant attribute feature map $\bm{X}_{re\_attr}$ and the identity feature map $\bm{X}_{ID}$. Finally, the ID-relevant attribute enhanced feature $\bm{f}_{re\_ID}$ are got according to the $\bm{X}_{re\_ID}$ with the spatial-temporal pooling. The ID-relevant attribute branch is formulated as following:
\begin{equation}
\begin{aligned}
    \bm{X}_{re\_ID}&=ASRE(\bm{X}_{re\_attr},F_{re}(\bm{X}_{ID})),
    \\
    \bm{f}_{re\_ID}&=W_{re}(F_{TP}(F_{SP}(\bm{X}_{re\_ID}))),
\label{eq:re_ID}
\end{aligned}
\end{equation}
where $ASRE$,  $F_{TP}$, $F_{SP}$, $W_{re}$, $F_{re}$ denote the proposed ASRE module, temporal pooling, spatial pooling, FC layer and 2d convolution with filter size of 1×1. $\bm{X}_{re\_ID} \in \mathbb{R}^{T \times \frac{C}{4} \times H \times W}$, $\bm{f}_{re\_ID} \in \mathbb{R}^{T \times \frac{C}{4}}$. The ID-relevant attribute feature and the corresponding prediction can be expressed as:
\begin{equation}
\begin{aligned}
    \bm{f}_{re\_attr}&=F_{TP}(F_{SP}(\bm{X}_{re\_attr})),
    \\
    \bm{p}_{re}&=W_1(\sigma(BN(W_2(\bm{f}_{re\_attr})))),
    \label{eq:re_attr}
\end{aligned}
\end{equation}
where $W_1$, $W_2$ are FC layers, $\sigma$ is the activation function and the BN is the batch norm layer. $\bm{f}_{re\_attr} \in \mathbb{R}^{\frac{C}{4}}$ is the ID-relevant attribute feature and $\bm{p}_{re} \in \mathbb{R}^{D_{re\_attr}}$ is the attribute prediction where $D_{re\_attr}$ is the number of ID-relevant attributes to predict. In this paper, we transform a multi-category classification problem into multiple binary-category classification problems. For example, there are five kinds of top color attribute on MARS dataset, we transform them into five binary classification problems. Therefore, $D_{re\_attr}$ equals to the sum of number of all categories for all attributes. 

The framework of ID-irrelevant attribute branch is same to the ID-relevant attribute branch. The ID-irrelevant attribute branch can be formulated as following:
\begin{equation}
\begin{aligned}
    \bm{X}_{ir\_ID}&=ASRE(\bm{X}_{ir\_attr},F_{ir}(\bm{X}_{ID})),
    \\
    \bm{f}_{ir\_ID}&=W_{ir}(F_{TP}(F_{SP}(\bm{X}_{ir\_ID}))),
\label{eq:ir_ID}
\end{aligned}
\end{equation}
where $F_{ir}$ and $W_{ir}$ are a 2d convolution layer with filter size of 1$\times$1 and FC layer. $\bm{X}_{ir\_ID} \in \mathbb{R}^{T \times \frac{C}{4} \times H \times W}$, $\bm{f}_{ir\_ID} \in \mathbb{R}^{T \times \frac{C}{4}}$. The ID-irrelevant attribute prediction can be expressed as:
\begin{equation}
\begin{aligned}
    \bm{f}_{ir\_attr}&=F_{TP}(F_{SP}(\bm{X}_{ir\_attr})),
    \\
    \bm{p}_{ir}&=W_3(\sigma(BN(W_4(\bm{f}_{ir\_attr})))),
\label{eq:ir_attr}
\end{aligned}
\end{equation}
where $W_3$, $W_4$ are FC layers.$\bm{f}_{ir\_attr} \in \mathbb{R}^{\frac{C}{4}}$ is the ID-irrelevant attribute feature and $\bm{p}_{ir} \in \mathbb{R}^{D_{ir\_attr}}$ is the ID-irrelevant attribute prediction where $D_{ir\_attr}$ is the number of ID-irrelevant attributes to predict.

In order to keep the dimension consistent, we also reduce the dimension of original identity feature map $\bm{X}_{ID}$ and calculate the corresponding feature $\bm{f}_{ID}$:
\begin{equation}
\begin{aligned}
    \bm{X}_{ID\_ID}&=F_{ID}(\bm{X}_{ID}),
    \\
    \bm{f}_{ID}&=F_{TP}(F_{SP}(\bm{X}_{ID\_ID})),
\label{eq:ID}
\end{aligned}
\end{equation}
where $F_{ID}$ is a 2d convolution layer with filter size of 1$\times$1, $\bm{X}_{ID\_ID} \in \mathbb{R}^{T \times \frac{C}{4} \times H \times W}$ and $\bm{f}_{ID} \in \mathbb{R}^{T \times \frac{C}{4}}$.

Combining Eq.~\eqref{eq:re_ID}, \eqref{eq:re_attr}, \eqref{eq:ir_ID}, \eqref{eq:ir_attr} and \eqref{eq:ID}, the final feature are calculated as: 
\begin{equation}
    \bm{f}_{final}=F_{fusion}(\bm{f}_{re\_attr},\bm{f}_{re\_ID},\bm{f}_{ID},\bm{f}_{ir\_attr},\bm{f}_{ir\_ID}),
\label{eq:final}
\end{equation}
where $\bm{f}_{final} \in \mathbb{R}^{c}$ and the $F_{fusion}$ is the fusion module which will be explained in Section~\ref{subsec:fuse}.

\subsection{Attribute Salient Region Enhance(ASRE) Module}

Background clutter often interferes in person Re-ID task, so some work are devoted to using attention for enhancing the features of human region~\cite{subramaniam2019co,chen2020deep}. However, their attention is calculated with the input image itself. Inspired by~\cite{chai2021semantically}, middle-level semantic labels of covariates often indicate the spatial locations of covariates. Therefore, additional middle-level semantic labels may indicate the spatial location of the person and make the attention more precise in person Re-ID. In this paper, While providing attribute information, the attribute annotation can also implicitly contain the relevant spatial cues of human body. Therefore the Attribute Salient Region Enhance (ASRE) module is proposed to find salient regions of pedestrian with attribute labels. 

The procedure of ID-relevant attribute branch and ID-irrelevant attribute branch is similar, so we explain the ASRE module of ID-relevant attribute branch as an example.  

Let's denote the $F_{re}(\bm{X}_{ID})$ as $\bm{X}_{I}$ and $\bm{X}_{re\_attr}$ as $\bm{X}_{a}$ which represent the identity feature map and the ID-relevant attribute feature map in Eq.~\eqref{eq:ID}, \eqref{eq:re_attr}. Self-attention is a powerful long-range relationship capture mechanism~\cite{vaswani2017attention}, we modified it into a cross-attention mechanism to capture the relationship between two feature maps. As shown in Fig.~\ref{fig:ASRE}, the Attribute-Identity cross attention map can be calculated as:
\begin{equation}
\begin{aligned}
    Att(\bm{X}_{I},\bm{X}_{A})=Softmax(\frac{\sigma({F_{Q}(\bm{X}_{I})})^{T} \sigma(F_{K}(\bm{X}_{A}))}{\sqrt{d_{k}}}),
\label{eq:cross attention}
\end{aligned}
\end{equation}
where $F_{Q}$ and $F_{K}$ are combination of convolution layers with filter size of 1$\times$1 and dimension merge operation. $\sigma$ is the activation function and $\sqrt{d_{k}}$ is the scaling factor to prevent excessive values. In practice, $\sigma$ is the ReLU function and $d_{k}$ equals to the feature dimension after the 1$\times$1 convolution. This attention map indicates the response degree of each pixel of the identity feature map to each pixel of the attribute feature map. Then the salient-region mask can be formulated as:
\begin{equation}
\begin{aligned}
    \bm{V}&=F_{V}(\bm{X}_{A}),
    \\
    \bm{X}_{V}&=OP_{split}(Att(\bm{X}_{I},\bm{X}_{A})\bm{V}),
    \\
    \bm{M}_{ASR}&=F_{mask}(BN(\bm{X}_{V})),
\label{eq:mask}
\end{aligned}
\end{equation}
where $F_{V}$ is combination of convolution layers with filter size of 1x1 and dimension merge operation, $OP_{split}$ is the operation of changing the dimension from $\frac{C}{4} \times HW$ to $\frac{C}{4} \times H \times W$ and $F_{mask}$ is a 2d convolution layer with filter size of 1$\times$1. $\bm{V} \in \mathbb{R}^{HW \times \frac{C}{4}}$, $\bm{X}_{V} \in \mathbb{R}^{\frac{C}{4} \times H \times W}$ and $\bm{M}_{ASR} \in \mathbb{R}^{H \times W}$

Combining with Eq. \eqref{eq:cross attention}, \eqref{eq:mask}, the attribute-enhanced feature map and final output of the ASRE module can be written as:
\begin{equation}
\begin{aligned}
\bm{X}_{enhanced}&=\bm{M}_{ASR} \otimes \bm{X}_{I},
\\
\bm{X}_{re\_ID}&=\bm{X}_{I}+\alpha*\bm{X}_{enhanced},
\label{eq:enhance}
\end{aligned}
\end{equation}
where $\otimes$ is \emph{Hadamard} product and $\alpha$ is a learnable parameter. The default value of $\alpha$ is 1.0 for both branches.

\subsection{Pose \& Motion-Invariant Triplet Loss (PMI Triplet Loss)}

The common triplet loss aims at enlarging the distance between positive and negative pairs. However, it can not narrow the intra-class distance caused by the variation introduced by pose and motion. Some works~\cite{zhu2020aware,chen2020deep,zheng2019pose} tried to solve this problem in image-based person Re-ID. In this work, with the obtained ID-irrelevant attribute feature $\bm{f}_{ir\_attr}$, we can measure the difference of pose and motion between positive sample pairs.

Considering that there are $K$ samples $\{\bm{I}_{1},\bm{I}_{2},...,\bm{I}_{K}\}$ of one person in a mini-batch and the ID-irrelevant attribute features of them are ${\{\bm{f}_{ir\_attr}^{1},\bm{f}_{ir\_attr}^{2},...,\bm{f}_{ir\_attr}^{K}\}}$ calculated in Eq.~\eqref{eq:ir_attr} . For every anchor sample $\bm{I}_{A}$, the triplet pairs are generated as following:
\begin{equation}
\begin{aligned}
AP&=\mathop{\arg\min}_{1\leq i \leq K, i\neq anchor}(||\bm{f}_{ir\_attr}^{i}-\bm{f}_{ir\_attr}^{A}||_2), 
\\
AN&=\mathop{\arg\max}_{1\leq i \leq K, i\neq anchor}(||\bm{f}_{ir\_attr}^{i}-\bm{f}_{ir\_attr}^{A}||_2),
\label{eq:hard example}
\end{aligned}
\end{equation}
and the triplet pair for anchor sample $I_{A}$ is $\{I_{A}, I_{AP}, I_{AN}\}$. Supposing there N triplets in a mini-batch which are denoted as $\{T_{i}|T_{i}=(\bm{f}_{final}^{A_{i}},\bm{f}_{final}^{AP_{i}},\bm{f}_{final}^{AN_{i}})\}$ where $\bm{f}_{final}$ is the final feature calculated in Eq. \eqref{eq:final}. Finally, the PMI triplet loss can be formulated as:
\begin{equation}
\begin{aligned}
\mathcal{L}_{PMI}&=\frac{1}{N}\sum^{N}_{i=1}max(d_i^--d_i^+,0),
\label{eq:PMI}
\end{aligned}
\end{equation}
where $N$ is the number of samples within a mini-batch and 
\begin{equation}
\begin{aligned}
d_i^- &= 1-\frac{\bm{f}_{final}^{A_i}\bm{f}_{final}^{AN_i}}{\left|\bm{f}_{final}^{A_i}\right|\left|\bm{f}_{final}^{AN_i} \right|},
\\
d_i^+ &= 1-\frac{\bm{f}_{final}^{A_i}\bm{f}_{final}^{AP_i}}{\left|\bm{f}_{final}^{A_i}\right|\left|\bm{f}_{final}^{AP_i} \right|},
\label{eq:cos}
\end{aligned}
\end{equation}
denote the distance of the final feature between ($I_{A}$, $I_{AN}$) and ($I_{A}$, $I_{AP}$). 

\subsection{Feature Fusion Strategy}
\label{subsec:fuse}

\begin{figure}[t]
    \centering
    \includegraphics[width=3.5in]{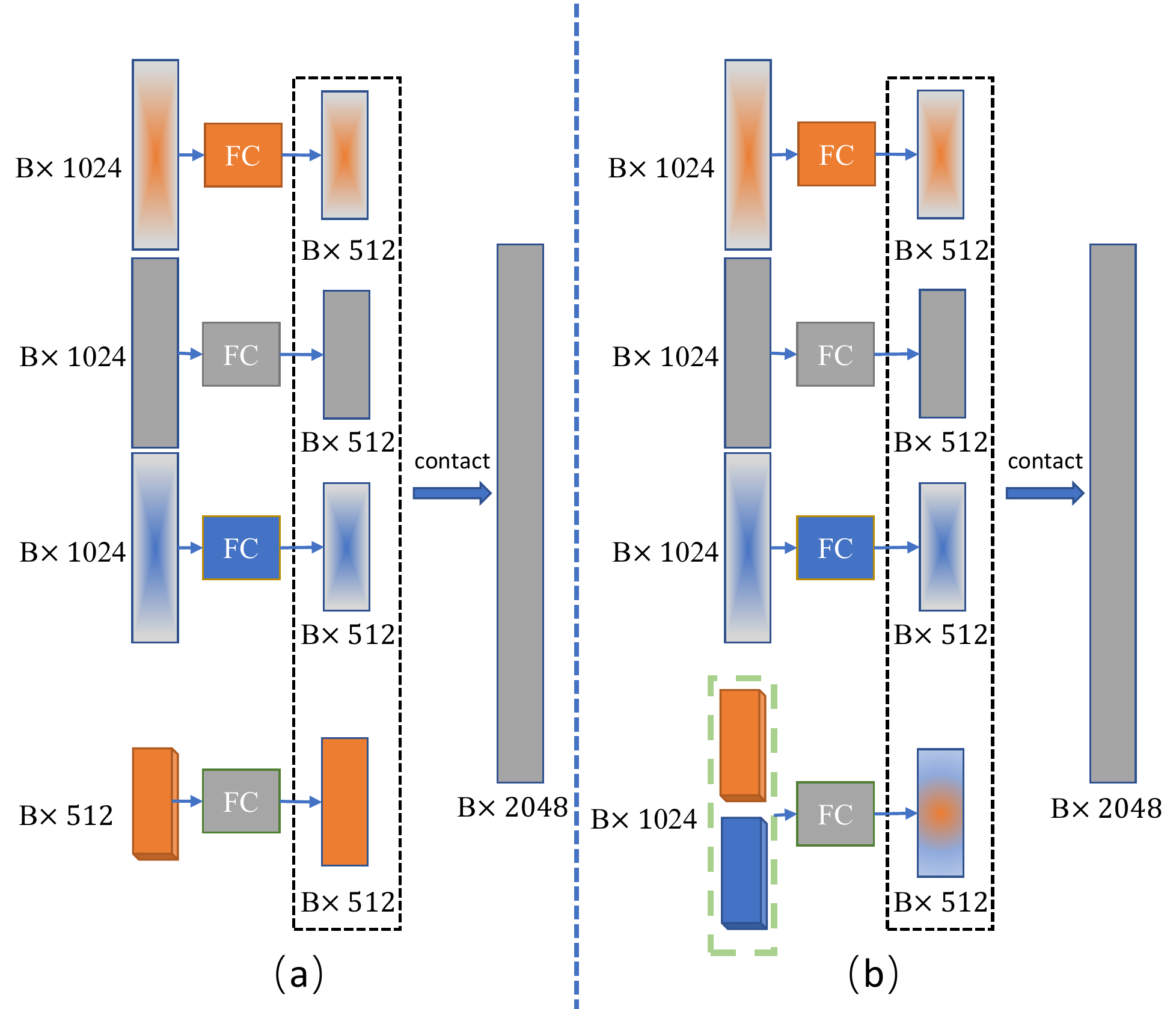}
    \setlength{\abovecaptionskip}{0pt}
    \setlength{\belowcaptionskip}{0pt}
    \caption{Fusion strategy of features of five branches. Each strip corresponds to the same one of Fig.~\ref{fig:network}. The orange-grey mixed strip, grey strip, blue-grey mixed strip, orange strip and blue strip denote ID-relevant enhanced identity feature, original identity feature, ID-irrelevant enhanced identity feature, ID-relevant attribute feature and ID-irrelevant attribute feature respectively. All attribute enhanced features go through a Fully connect layer to reduce dimensions. In fusion strategy (a), only ID-relevant attribute feature are fused into final feature while in fusion strategy (b) we do a early fusion of ID-relevant attribute feature and ID-irrelevant attribute feature.}
    \label{fig:fusion}
    \vspace{-0.5cm} 
\end{figure}

As shown in Fig.~\ref{fig:ASRE}, in the end of network pipeline, features from five branches need to be fused into one identity feature. The proposed two fusion strategies are shown in Fig.~\ref{fig:fusion}. For both strategies, $\bm{f}_{re\_ID}, \bm{f}_{ID}, \bm{f}_{ir\_ID}$ need to go through a FC layer for dimension reduction. For strategy (a), only the feature of ID-relevant attributes $f_{re\_ID}$ are fused in the final feature. For strategy (b), we contact features of ID-relevant and ID-irrelevant attributes into one attribute feature. Then the attribute goes through a FC layer and is contacted with other features.

Generally, the feature of ID-irrelevant is considered harmful to the Re-ID task. However, we argue that the ID-irrelevant features including pose and motion information can help to filter ID-relevant features. The experiments confirmed our supposition.

\subsection{The Overall Loss}

As illustrated in Fig.~\ref{fig:ASRE}, our losses consist of five parts, i.e. weighted regularization triplet loss, center loss, cross entropy loss, binary cross entropy loss and the pose\&motion-invariant loss. 

Weighted regularization triplet loss was proposed in~\cite{ye2021deep} which can be expressed as:
\begin{equation}
\begin{aligned}
\mathcal{L}_{wrt}=\sum_{i=1}^Nlog(1+exp(\sum_j(w_{ij}^pd_{ij}^p)-\sum_k(w_{ik}^nd_{ik}^n))
\\
w_{ij}^p=\frac{exp(d_{ij}^p)}{\sum_{d_{ij}^p\in\mathcal{P}_i}exp(d_{ij}^p)},
w_{ik}^n=\frac{exp(-d_{ik}^n)}{\sum_{d_{ik}^n\in\mathcal{N}_i}exp(-d_{ik}^n)}
\label{eq:weight trip}
\end{aligned}
\end{equation}
where $(i,j,k)$ represents a triplet within each training batch. For anchor $i$, $\mathcal{P}_{i}$ is the corresponding positive set and $\mathcal{N}_{i}$ is the negative set. $d_{ij}^p/d_{ik}^n$represents the pairwise distance of a positive/negative sample pair.

Center loss was proposed in \cite{wen2016discriminative} which aims at minimizing the intra-class variations. It can be formulated as:
\begin{equation}
\mathcal{L}_{cent}=\frac{1}{2}\sum_{i=1}^N\|\bm{f}_{i}-\bm{c}_{y_i}\|_2,
\label{eq:center}
\end{equation}
where $\bm{c}_{yi} \in \mathbb{R}^{c}$ denotes the center of $yi$th person's features and is a learnable parameter.   

Cross entropy loss has been used in Re-ID task widely. In this paper we used the cross entropy loss with label smooth as ~\cite{luo2019bag}. Supposing there are $M$ persons in the training dataset, for the obtained final feature $\bm{f}_{final}$ in Eq.\eqref{eq:final}, we calculated the prediction probability of sample $i$ as:
\begin{equation}
    \bm{p}_i=W_{ID}(BN(\bm{f}_{final,i})),
\end{equation}
where BN is a batchnorm layer, $W_{ID}$ is a FC layer and $\bm{p}_i \in \mathbb{R}^M$.  
Then the cross entropy loss with label smooth can be expressed as:
\begin{equation}
\begin{aligned}
\mathcal{L}_{xent}=\sum_{i=1}^N\sum_{j=1}^M-q_{ij}log(\bm{p}_{ij})\left\{  
           \begin{aligned}
            q_{ij} & = 1-\frac{N-1}{N}\epsilon, y_i=j \\
            q_{ij} & = \frac{\epsilon}{N}, otherwise
            \end{aligned}
            \right.,
\label{eq:xent}
\end{aligned}
\end{equation}
where $y_i$ is the ID label of sample $i$.

The binary cross entropy loss is used for attribute prediction. We merge the obtained attribute prediction probability $\bm{p}_{re}$ and $\bm{p}_{ir}$ in Eq.~\eqref{eq:re_attr}, \eqref{eq:ir_attr} and denote them as $\bm{p}_{attr}$. The $i$th prediction of sample $i$ is denoted as $\bm{p}^{attr}_i \in \mathbb{R}^{D_{attr}}$ where $D_{attr}=D_{re\_attr}+D_{ir\_attr}$. Then the binary cross entropy loss of attributes are formulated as:
\begin{equation}
\begin{aligned}
\mathcal{L}_{BCE}=\sum_{i=1}^N\sum_{j=1}^{D_{attr}}-y^{attr}_{ij}log(\bm{p}^{attr}_{ij}),
\label{eq:bce}
\end{aligned}
\end{equation}
where $y^{attr}_{ij} \in \{0,1\}$ is the one-hot label of $j$th attribute of $i$th sample.

Finally, combining Eq.~\eqref{eq:PMI}, \eqref{eq:weight trip}, \eqref{eq:center}, \eqref{eq:xent}, \eqref{eq:bce}, the loss function of the model can be written as:
\begin{equation}
    \mathcal{L}=\mathcal{L}_{xent}+\mathcal{L}_{wrt}+\lambda_{cent}\mathcal{L}_{cent}+\lambda_{BCE}\mathcal{L}_{BCE}+\lambda_{PMI}\mathcal{L}_{PMI},
\end{equation}
where $\lambda_{cent}$, $\lambda_{BCE}$, $\lambda_{PMI}$ are hyper-parameters.

\section{Experiment}
\label{sec:exp}

\subsection{Datasets and Evaluation Protocols}
We evaluate the proposed model on two large-scale video person Re-ID datasets: DukeMTMC-VideoReID~\cite{ristani2016performance} and MARS~\cite{zheng2016mars}. DukeMTMC-VideoReID comprises around 4,832 videos from 1,812 identities and we use \emph{Duke} for abbreviation. There are 702 and 702 identities for training and testing in Duke as well as 408 identities for distraction. The bounding boxes are annotated manually. MARS is the largest dataset for video-based person Re-ID and has 17,503 sequences of 1,261 identities  and 3,248 dis-tractor sequences. Among 1,261 identities, 625 identities are for training and the other 636 identities are for testing. The annotation of attributes we use is provided by Chen et al.~\cite{chen2019temporal}.   


Cumulative Matching Characteristic (CMC) is widely used to evaluate the performance of person Re-ID models and we adopt the CMC to evaluate our model. Also, the mean Average Precision (mAP) is also adopted for evaluation.

\subsection{Implementation Details}

\begin{table}[t]
\caption{Performance(\%) comparison with related works on MARS. ``a'' and ``b'' denote the fusion strategy. PMI means the model is trained with PMI triplet loss. ``Mixing'' means the result is tested under the situation of mixing the query set and the gallery set as the new gallery set. AP3D and TCLNet are tested under this special situation according to their code to avoid the lack of corresponding gallery samples. The highest and second highest performance are \textbf{bold} and \underline{underlined} under usual setup. The highest performance are \textbf{bold} under Mixing setup.} 
\label{tab:mars}
\begin{tabular}{c|c|c|c|c|c}
\hline
Setup                 & Method                                                   & mAP                               & Rank-1                            & Rank-5                            & Rank-20                        \\ \hline
\multirow{17}{*}{Usual} & BoW+kissme~\cite{zheng2016mars}    & 15.5                              & 30.6                              & 46.2                              & 59.2                           \\
                              & IDE+XQDA ~\cite{zheng2016mars}     & 47.6                              & 65.3                              & 82.0                              & 89.0                           \\
                              & Wu et al.~\cite{wu2018and}  & - & 73.5 & 85.0  & 97.5   \\
                              & MG-TCN~\cite{li2020hierarchical}  & 77.7 & 87.0 & 95.1 & 98.2 \\
                              & STA~\cite{fu2019sta}               & 80.8                              & 86.3                              & 95.7                              & 98.1                           \\
                              & Attribute~\cite{zhao2019attribute} & 78.2                              & 87.0                              & 95.8                              & \underline{98.7} \\
                              & HMN~\cite{wang2021robust}  & 82.6 & 88.5 
                              & 96.2 &98.1 \\
                              & VRSTC~\cite{hou2019vrstc}          & 82.3                              & 88.5                              & 96.5                              & 97.4                           \\
                              & GLTR~\cite{li2019global}           & 78.5                              & 87.0                              & 95.8                              & 98.2                           \\
                              & COSAM~\cite{subramaniam2019co}     & 79.9                              & 84.9                              & 95.5                              & 97.9                           \\
                              & AMEM~\cite{li2020appearance}       & 79.3                              & 86.7                              & 94.0                              & 97.1                           \\
                              & MGRA~\cite{zhang2020multi}         & 85.9                              & 88.8                              & \underline{97.0} & 98.5                           \\
                              & STGCN~\cite{yang2020spatial}       & 83.7                              & 89.9                              & 96.4                              & 98.3                           \\
                              & AFA~\cite{chen2020temporal}        & 82.9                              & 90.2 & 96.6                              & -                              \\
                              & MGH~\cite{yan2020learning}         & 85.8                              & 90.0                              & 96.7                              & 98.5                           \\
                              & ASTA~\cite{zhu2020asta}     & 84.1   & \textbf{90.4}     & 97.0   & \textbf{98.8}               \\
                              & TALNet~\cite{liu2020temporal}      & 82.3                              & 89.1                              & 96.1                              & 98.5                           \\
                              & PhD~\cite{Zhao_2021_CVPR}       &    86.2      &   88.9       & \underline{97.0}      & 98.6                 \\
                              & ASANet-b(Ours)                                           & \textbf{86.6}    & \underline{90.3}    & 96.8                              & 98.4                           \\
                              & ASANet-a-PMI(Ours)                                       & \underline{86.3} & 89.5                              & \textbf{97.1}    & \underline{98.7} \\ \hline
\multirow{4}{*}{Mixing} & TCLNet~\cite{hou2020temporal}      & 85.1                              & 89.8                              & -                                 & -                              \\
                              & AP3D~\cite{gu2020appearance}       & 85.6                              & 90.7                              & -                                 & -                              \\
                              & BiCnet~\cite{hou2021bicnet}           & \textbf{86.0}     & 90.2     & -    & -                            \\
                              & ASANet-b(Ours)                                           & \textbf{86.0}    & \textbf{91.1}    & 97.0                              & 98.4                           \\
                              & ASANet-a-PMI(Ours)                                       & \textbf{86.0}    & 90.6                              & \textbf{97.1}    & \textbf{98.7} \\ \hline
\end{tabular}
\end{table}

The baseline of our net is AGW-Net~\cite{ye2021deep} which is based on ResNet-50~\cite{he2016deep} pre-trained on ImageNet~\cite{deng2009imagenet}. The backbone part in Fig.~\ref{fig:network} only contains the first four layers of ResNet-50. As the same as~\cite{luo2019bag}, each image is resized to 256$\times$128, random horizontal flip, random crop and random erasing are adopted as data augmentation. We applied the constrain random sampling strategy~\cite{li2018diversity} to randomly sample $T=6$ frames from every clip. We train our network for 700 epochs in total, with the initial learning rate of 0.0003 and decayed it at the 100, 250, 350 epoch. For parameters of the ASA-network, we use Adam~\cite{kingma2014adam} algorithm with weight decay of 5e-4 and for parameters of the center loss in Equation~\eqref{eq:center}, we use SGD~\cite{bottou2010large} algorithm with learning rate of 0.5. For every mini-batch, we sample eight identities, each with four tracklets, to form a mini-batch of size 8$\times$4$\times$6=192 images. $\lambda_{cent}$, $\lambda_{BCE}$ are set to 1.5 and 0.0005. $\lambda_{PMI}$ is set to 0.005 initially and turn to 0.01 when $\mathcal{L}_{BCE}$ is smaller than 0.15 which means the prediction of pose and motion is accurate enough. Our framework is implemented with Pytorch toolbox and two NVIDIA GTX 1080Ti GPUs.

\begin{table}[]
\caption{Performance(\%) comparison with related works on Duke. 'b' denote the fusion strategy. 'PMI' means the model is trained with PMI triplet loss.} 
\label{tab:duke}
\begin{center}
\begin{tabular}{c|c|c|c|c}
\hline
Method             & mAP   & Rank-1 & Rank-5 & Rank-20 \\ \hline
EUG~\cite{wu2018exploit}                & 78.3  & 83.6   & 94.6   & 97.6    \\
ETAP-Net~\cite{wu2018exploit}           & 78.3  & 83.6   & 94.6   & 97.6    \\
STE-NVAN~\cite{liu2019spatially}           & 93.5  & 95.2   & -      & -       \\
VRSTC~\cite{hou2019vrstc}              & 93.5  & 95.0   & 99.1   & -       \\
STA~\cite{fu2019sta}                & 94.9  & 96.2   & 99.3   & 99.6    \\
Wu et al.~\cite{wu2020adaptive}          & 94.2  & 96.7   & 99.2   & 99.7    \\
GLTR~\cite{li2019global}               & 93.74 & 96.29  & 99.3   & 99.7    \\
HMN~\cite{wang2021robust}            & 95.1 & 96.3  & 99.2 & 99.8 \\
STGCN~\cite{yang2020spatial}            & 95.7 &\underline{97.3}   & 99.3   & 99.7 \\ 
AP3D~\cite{gu2020appearance}            & 96.1 & 97.2   &  -     & - \\
TCLNet~\cite{hou2020temporal}           & \underline{96.2} & 96.9  &  -      & -\\
ASANet-b(Ours)     & \textbf{97.1}  & \underline{97.3}   & \textbf{99.9}   & \textbf{100.0}    \\
ASANet-b-PMI(Ours) & \textbf{97.1}  & \textbf{97.6}   & \textbf{99.9}   & \textbf{100.0}    \\ \hline
\end{tabular}
\end{center}
\vspace{-0.5cm}
\end{table}

\begin{table}[t]
\begin{center}
\caption{Cross-dataset evaluations between MARS and DukeMTMC-Video (Duke for short)(\%).}
\label{tab:cross}
\begin{tabular}{c|c|c|c|ccc}
\hline
\multirow{2}{*}{Method} & \multicolumn{3}{c|}{MARS-\textgreater{}Duke}  & \multicolumn{3}{c}{Duke-\textgreater{}MARS}                                             \\ \cline{2-7} 
& mAP   & Rank-1        & Rank-5    & \multicolumn{1}{c|}{mAP}  & \multicolumn{1}{c|}{Rank-1}        & Rank-5        \\ \hline
QAN~\cite{chen2020temporal}                     & 33.0          & 38.9          & 59.3          & \multicolumn{1}{c|}{24.1}          & \multicolumn{1}{c|}{43.3}          & 58.8          \\
AFA~\cite{chen2020temporal}                     & 34.6          & 41.5          & 59.1          & \multicolumn{1}{c|}{24.5}          & \multicolumn{1}{c|}{44.2}          & 58.8          \\
Ours                    & \textbf{57.8} & \textbf{57.8} & \textbf{77.9} & \multicolumn{1}{c|}{\textbf{35.1}} & \multicolumn{1}{c|}{\textbf{58.0}} & \textbf{69.7} \\ \hline
\end{tabular}
\end{center}
\vspace{-0.5cm}
\end{table}

\begin{table}[t]
\begin{center}
\caption{Ablation Study of different branches on MARS(\%).} 
\label{tab:branches}
\begin{tabular}{ccc|c|c|c|c}
\hline
\multicolumn{3}{c|}{Branches}          & \multicolumn{4}{c}{Metrics}     \\ \hline
Identity & ID-re.       & ID-ir.        & mAP  & rank-1 & rank-5 & rank-20 \\ \hline
\checkmark        &             &               & 83.9 & 89.2   & 96.8   & 98.2        \\
         & \checkmark           &               & 85.1 & 89.8   & 96.9   & 98.5    \\
         &             & \checkmark             & 84.8 & 89.8   & 96.8   & 98.4    \\
\checkmark        & \checkmark           &               & 85.3 & 90.3   & 96.6   & 98.2    \\
\checkmark        &             & \checkmark             & 84.6 & 89.5   & 96.5   & 98.3    \\
         & \checkmark           & \checkmark             & 85.8 & 89.9   & 96.8   & 98.3    \\ \hline
\checkmark        & \checkmark           & \checkmark             & \textbf{85.8} & \textbf{90.3}   & \textbf{96.9}   & \textbf{98.5}    \\ \hline
\end{tabular}
\end{center}
\vspace{-0.5cm}
\end{table}

\subsection{Comparison with the State-of-the-art Approaches}

In this section, we compare the proposed method with state-of-the-art methods on two video person Re-ID benchmarks. 

The results on MARS dataset are reported in Table~\ref{tab:mars}. There are 140 samples in the query set lack the corresponding samples in gallery set on MARS dataset. The usual testing setup is to remove these 140 samples from the query set. However, we find that Gu et al.~\cite{gu2020appearance} and Hou et al.~\cite{hou2020temporal} merge the query set and gallery set to form the new gallery set to avoid this problem according to their codes which may make the rank-1 value higher and the mAP value lower. We compare the results under two testing setups separately for fair. Some methods don't make their code available nor tell the detail in their papers, we make a reasonable guess that they tested their methods under the usual setup. Our experiments show that different fusion strategies and whether to use PMI triplet loss lead to different result on CMC. So we list the two combinations with the highest mAP here. The more analysis of fusion strategies and PMI triplet loss will be presented in the subsection of ablation study. 

Under the normal setup, the mAP and Rank-1 accuracy of our method with fusion strategy (b) and without PMI triplet loss (ASANet-b) is 86.6\% and 90.3\% which have obvious improvement compared with SOTA methods. Especailly, compared with the first work of using attribute \cite{zhao2019attribute}, ASANet-b improves 8.4\% on mAP and 3.3\% on Rank-1. AMEM~\cite{li2020appearance} considers both ID-relevant attributes and ID-irrelevant attributes as we do, we outperform AMEM~\cite{li2020appearance} by 7.3\% on mAP and 3.6\% on Rank-1. The Rank-1 value of our ASANet-b is close to ASTA~\cite{zhu2020asta} which also put forward the 'motion' concept, but the mAP value of our method is 2.5\% higher than it. To the same as our method, TALNet~\cite{liu2020temporal} uses the more accurate annotated attribute label and our ASANet-b is also 4.3\% and 1.2\% ahead of it on mAP and Rank-1. The mAP and Rank-1 value of our method with fusion strategy (a) and with PMI triplet loss(ASANet-a-PMI) is a little lower than ASANet-b. But ASANet-a-PMI achieves the highest rank-5 and rank-20 score which is more important in practical application.

Under the mixing setup, the mAP value of both ASANet-b and ASANet-a-PMI is higher than other SOTA methods and the Rank-1 of ASANet-b also outperforms other methods.

The results on Duke dataset are reported in Table~\ref{tab:duke}. Our methods outperform the SOTA method TCLNet~\cite{hou2020temporal} by 0.9\% on mAP which is a significant improvement since the mAP value is pretty high. It's worth noting that our method achieve 100\% on rank-20 which indicates that our method can achieve 100\% success in practical video-based person re-ID task (For example, tracking of escaped criminal) with manual screening.
\vspace{-0.3cm}

\begin{table}[t]
\begin{center}
\caption{Ablation Study of fusion strategies, ASRE module and PMI triplet loss on MARS(\%).} 
\label{tab:ASRE}
\begin{tabular}{ccc|c|c|c|c}

\hline
Fusion strategy & ASRE & PMI & mAP           & rank-1        & rank-5        & rank-20       \\ \hline
a               &      &     & 85.1          & 89.9          & 96.5          & 98.5          \\
a               &      & \checkmark   & 85.5          & 90.2          & 97.0          & 98.5          \\
a               & \checkmark    &     & 85.3          & 90.4          & 97.1          & 98.5          \\
a               & \checkmark    & \checkmark   & \textbf{86.0} & \textbf{90.6} & \textbf{97.1} & \textbf{98.7} \\ \hline
b               &      &     & 85.4          & 90.4          & 96.9          & 98.5          \\
b               &      & \checkmark   & 85.7          & 90.5          & \textbf{97.3} & 98.5 \\
b               & \checkmark    &     & \textbf{86.0} & \textbf{91.1} & 97.0          & 98.4          \\
b               & \checkmark    & \checkmark   & 85.8          & 90.8          & 97.0          & \textbf{98.6}          \\ \hline
\end{tabular}
\end{center}
\vspace{-0.4cm}
\end{table}

\begin{table}[!t]
\begin{center}
\caption{Ablation Study of fusion strategies and PMI triplet loss on Duke(\%).} 
\label{tab:duke_pmi}
\begin{tabular}{cc|c|c|c|c}
\hline
Fusion strategy & PMI & mAP           & rank-1        & rank-5        & rank-20        \\ \hline
a               &     & 96.7          & 97.0          & 99.7          & 100.0          \\
a               & \checkmark   & 97.1          & 97.4          & 99.9          & 100.0          \\
b               &     & 97.1          & 97.3          & 99.9          & 100.0          \\
b               & \checkmark   & \textbf{97.1} & \textbf{97.6} & \textbf{99.9} & \textbf{100.0} \\ \hline
\end{tabular}
\end{center}
\vspace{-0.3cm}
\end{table}

\begin{table}[t]
\begin{center}
\caption{Ablation study of attribute supervision(\%).}
\label{tab:bce}
\begin{tabular}{c|c|c|c|c}
\hline
Method                 & mAP  & Rank-1 & Rank-5 & Rank-20 \\ \hline
ASANet-b w/o BCE loss  & 85.5 & 89.8   & 97.0   & 98.6    \\
ASANet-b with BCE loss & 86.0 & 91.1   & 97.0   & 98.4    \\ \hline
\end{tabular}
\end{center}
\vspace{-0.3cm}
\end{table}

\subsection{Cross-dataset Evaluation}

In real surveillance systems, due to the limitation of data annotation, models should be able to cope with data from different sources. Especially, our method has achieved excellent performance on Duke dataset. In order to verify that our method is not over fitting on Duke, we do the cross-dataset experiments. The results are shown in Table~\ref{tab:cross}. We use ASANet-b and ASANet-a-PMI, which have the best performance on MARS and Duke themselves to do the cross-dataset experiments. We also compare the results with the methods proposed by Chen et al.~\cite{chen2020temporal} and our methods show obvious advantages.

%

\begin{figure*}[t]
    \centering
    \includegraphics[width=6.8in]{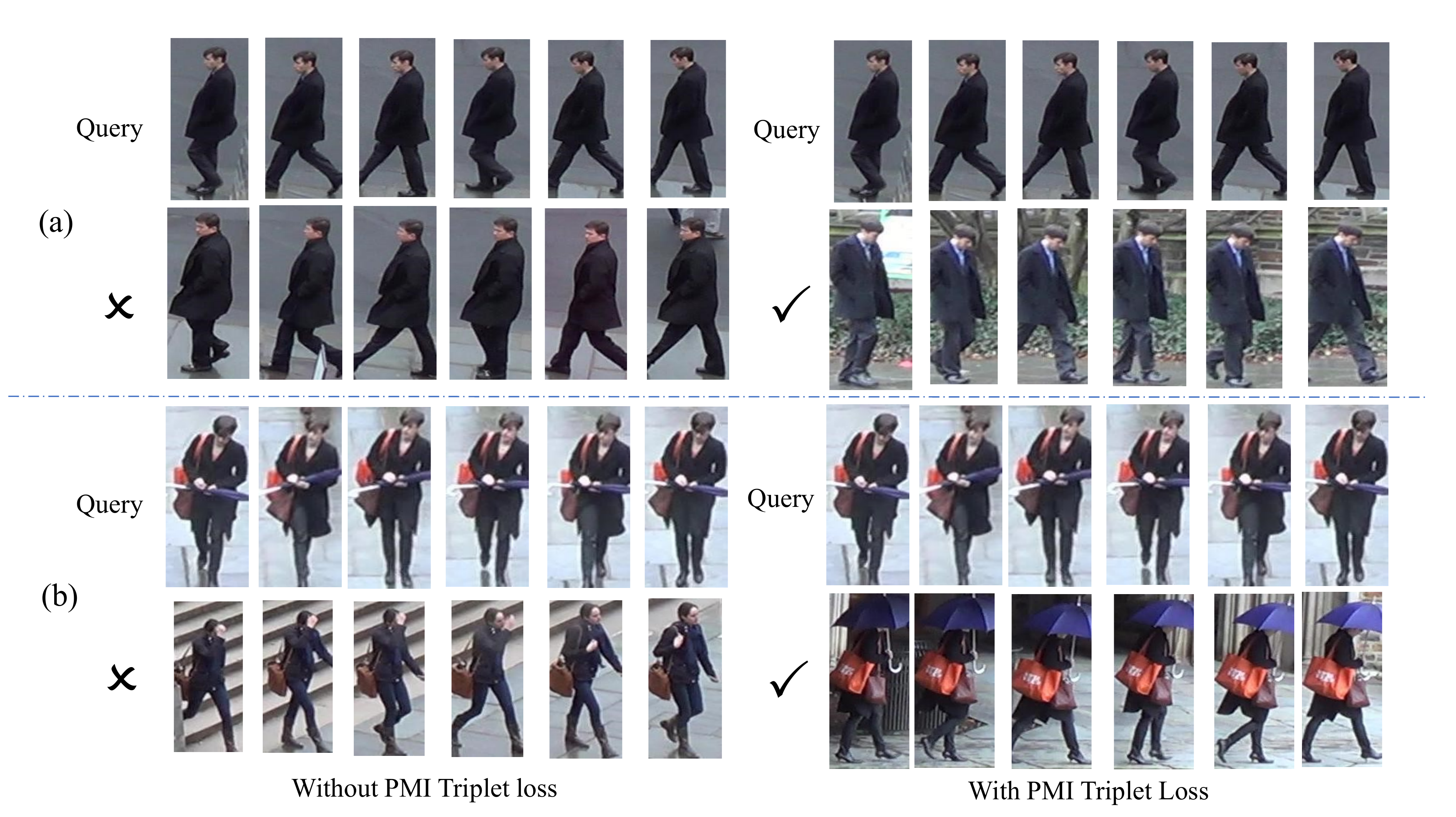}
    \setlength{\abovecaptionskip}{0pt}
    \setlength{\belowcaptionskip}{0pt}
    \caption{Top 1 retrieval results of the ASANet-b and ASANet-b-PMI. In exemplar sequence (a), the retrieved sequence without PMI triplet loss and the query sequence have almost the same pose and action, and their clothes and background are also extremely similar. But it can be seen from the head that they are not the same person. Although the sequences retrieved with PMI triplet loss are different in pose and background, they are the same person. Sequence (b) shows that our model is also robust to pose change caused by camera view difference.}
    \label{fig:PMI}
    \vspace{-0.4cm} 
\end{figure*}

\subsection{Ablation Study}

In this subsection, we conduct experiments to verify the effectiveness of the proposed methods. Experiments are mainly conducted on MARS dataset since it has the largest amount of sequences. Also, our experiments are done under the mixing setup.

\subsubsection{Effects of each branch}
The result in Table~\ref{tab:branches} shows the influence of each branch. ID-re. and ID-ir. denote the ID-relevant attribute enhanced feature and the ID-irrelevant attribute enhanced feature respectively. Note that we only use the enhanced feature and don't fuse attribute features here. Also, all the experiments in Table~\ref{tab:branches} are done with ASRE module and PMI triplet loss. All of the branches make sense. It can be seen that the Identity branch and ID-re branch plays a more important role in improving the performance which is congenial with common sense.

\subsubsection{Effects of the fusion strategies, ASRE module and the PMI triplet loss}

On the basis of the three-branch framework, we explore the role of the fusion strategies, the ASRE module and the PMI triplet loss. The results are reported in Table~\ref{tab:ASRE}. It can be seen that whether under the fusion strategy (a) or under the fusion strategy (b), ASRE module or PMI triplet loss can boost the performance. Under the fusion strategy a, the joint use of ASRE module and PMI triplet loss can further improve the performance. On the contrary, the joint use of ASRE module and PMI triplet loss will decrease the performance of the model under the fusion strategy (b), In order to explore the reason of this phenomenon, we observe the bad cases which is identified successfully with adding ASRE module only and failed with adding both ASRE and PMI. All of these bad cases retrieved the dis-tractor sequences which contain no person. In the training set, we have no dis-tractor sequences and the ASANet-b-PMI maybe is too complex to generalize to dis-tractor sequences which have no person.

To further prove the effectiveness of our proposed PMI triplet loss, we do the ablation study of PMI triplet loss on the Duke dataset which doesn't have sequences without pedestrian. The results are shown in Table~\ref{tab:duke_pmi}.

\subsubsection{Effects of the Attribute Assistance}

In order to explore whether attribute plays a role or whether our framework itself plays a role, we do the ablation experiments on the BCE loss. The results are shown in Table~\ref{tab:bce}. It can be seen that the attribute supervision can make significant improvement of mAP and Rank-1 value.

\begin{figure*}[!h]
    \centering
    \includegraphics[width=7in]{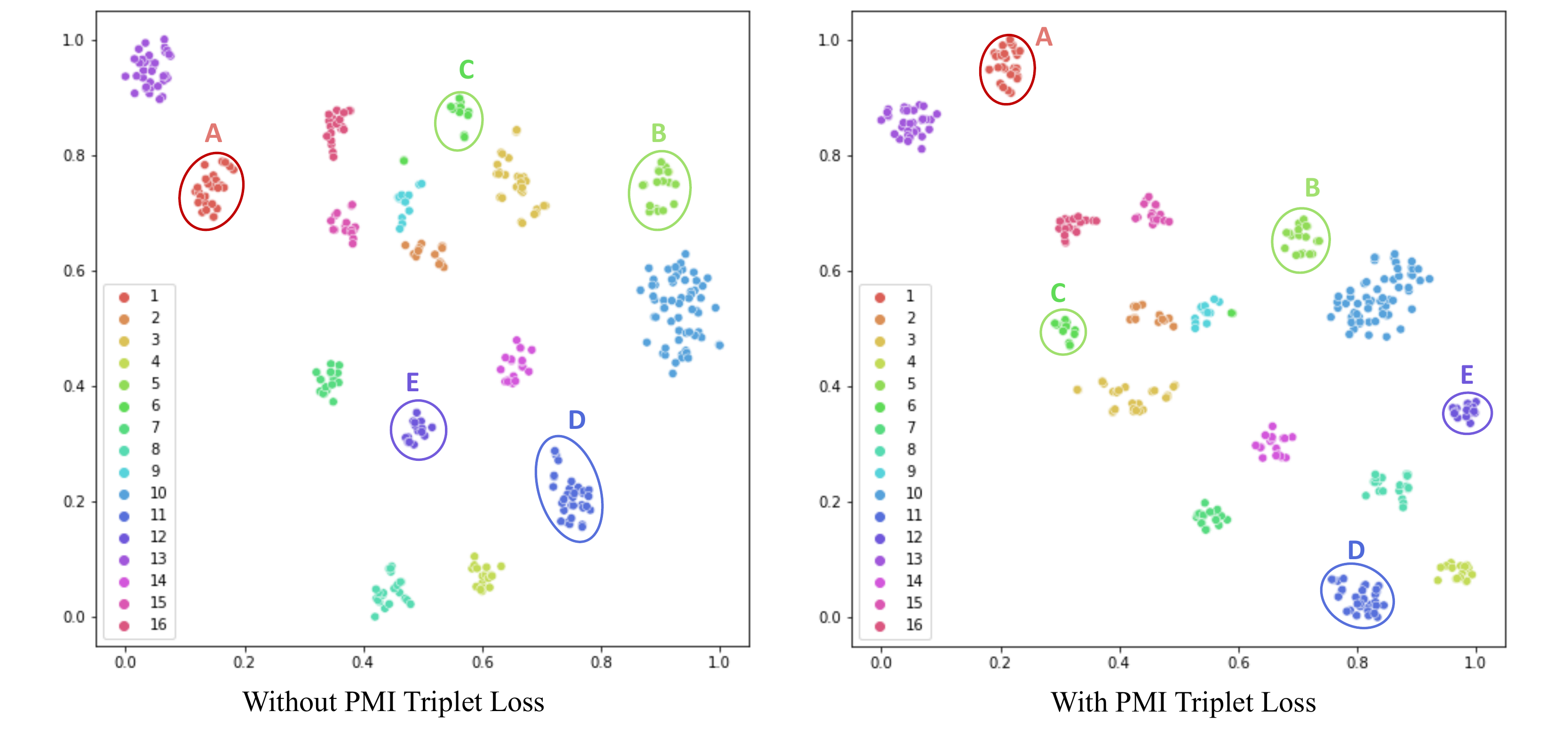}
    \setlength{\abovecaptionskip}{0pt}
    \setlength{\belowcaptionskip}{0pt}
    \caption{Visualization of feature distribution of the first 16 subjects on MARS gallery dataset using t-SNE. Subjects whose numbers of samples less than 10 is skipped because they are unsuitable for observation. It can be seen that the proposed PMI triplet loss can narrow the intra-class distance to a certain extent. For example, features of subject A, D, E can be divided into more sub-classes without PMI triplet loss than with PMI triplet loss.}
    \label{fig:PMI_tsne}
    \vspace{-0.4cm} 
\end{figure*}

\begin{figure*}[!h]
    \centering
    \includegraphics[width=7in]{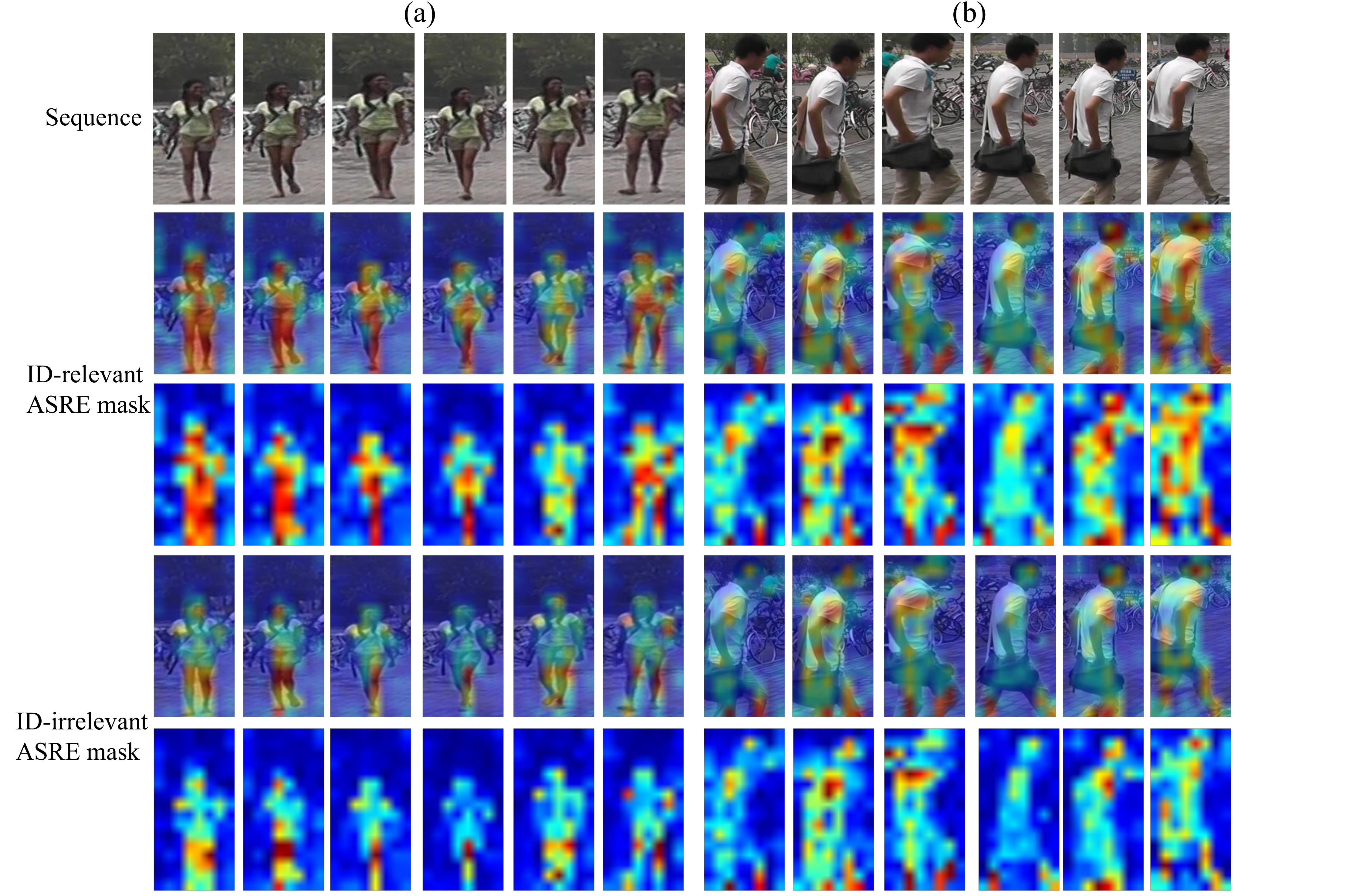}
    \setlength{\abovecaptionskip}{0pt}
    \setlength{\belowcaptionskip}{0pt}
    \caption{Attention mask generated by ASRE module. It can be seen that the mask regions generated by the two branches are roughly the same and can accurately cover the human body regions. But they differs in details. ID-relevant ASRE mask pay more attention to the detail of all the human body while ID-irrelevant ASRE mask only pay more attention to legs and joints of human body. ID-irrelevant ASRE mask can outline the edge of human body and reduce the influence of background more accurately.}
    \label{fig:exapmle_mars}
    \vspace{-0.4cm} 
\end{figure*}

\subsection{Visualization}

In order to illustrate the effectiveness of our method, we do some visualization.

Firstly, we visualize the effect of PMI triplet loss. According to the results on Duke dataset shown in Table~\ref{tab:duke_pmi}, the improvement caused by PMI triplet loss is mainly reflected in rank-1 value. Therefore, we show some top-1 retrieval results in Fig.~\ref{fig:PMI}. It can be seen that PMI triplet loss can cope with the change of pose and motion perfectly. Especially in Fig.~\ref{fig:PMI} (a), the retrieved sequence share almost the same pose and motion with the query sequence. Also, we choose the first 16 subjects (ordered by ID number) on MARS gallery dataset to visualize the feature distribution by t-SNE~\cite{van2008visualizing}. Subjects whose numbers of samples less than 10 is skipped because they are unsuitable for observation The results are shown in Fig.~\ref{fig:PMI_tsne}. It can be seen that PMI triplet loss can effectively narrow the intra-class distance. 

Secondly, to explore the efficacy of proposed ASRE module, we visualized the mask generated by the ASRE module. As illustrated in Fig.~\ref{fig:exapmle_mars}, the generated mask can mainly focus on the pedestrian body region and pay more attention to the area which has salient details. It's consistent with our design that mask generated by ID-relevant branch pay more attention to the region of human ID-relevant attributes such as the shoulder bag in Fig.~\ref{fig:exapmle_mars} (b) and mask generated by ID-irrelevant branch can outline the edge of human body more accurately and reduce the influence of background such as Fig.~\ref{fig:exapmle_mars} (a).

\section{Conclusion}
\label{sec:conclude}

In this work, we propose a comprehensive study on attribute-assisted video person re-identification. We introduced both ID-relevant attributes and ID-irrelevant attributes. To integrates attribute information into the Re-ID task we propose the ASA-Net and PMI triplet loss. The former predicts attributes, extract identity features at the same time and further enhance them by the attribute-salient regions learned from the ASRE module. With the extracted ID-irrelevant attribute feature, we form the PMI triplet loss to narrow the intra-class distance caused by the change of pose and motion. We conducted experiments on two challenging datasets and outperforms the state-of-the-art methods. Some visualizations are also presented to illustrate the effectiveness of our method.


%





\ifCLASSOPTIONcaptionsoff
  \newpage
\fi



%
\bibliographystyle{IEEEtran}
\bibliography{ref}

\begin{thebibliography}{10}

\bibitem{liu2012person}
Chunxiao Liu, Shaogang Gong, Chen~Change Loy, and Xinggang Lin,
\newblock ``Person re-identification: What features are important?,''
\newblock in {\em ECCV}. Springer, 2012, pp. 391--401.

\bibitem{kviatkovsky2012color}
Igor Kviatkovsky, Amit Adam, and Ehud Rivlin,
\newblock ``Color invariants for person reidentification,''
\newblock {\em IEEE TPAMI}, vol. 35, no. 7, pp. 1622--1634, 2012.

\bibitem{zhao2017deeply}
Liming Zhao, Xi~Li, Yueting Zhuang, and Jingdong Wang,
\newblock ``Deeply-learned part-aligned representations for person
  re-identification,''
\newblock in {\em ICCV}, 2017, pp. 3219--3228.

\bibitem{hirzer2011person}
Martin Hirzer, Csaba Beleznai, Peter~M Roth, and Horst Bischof,
\newblock ``Person re-identification by descriptive and discriminative
  classification,''
\newblock in {\em SCIA}. Springer, 2011, pp. 91--102.

\bibitem{bai2017scalable}
Song Bai, Xiang Bai, and Qi~Tian,
\newblock ``Scalable person re-identification on supervised smoothed
  manifold,''
\newblock in {\em CVPR}, 2017, pp. 2530--2539.

\bibitem{xiong2014person}
Fei Xiong, Mengran Gou, Octavia Camps, and Mario Sznaier,
\newblock ``Person re-identification using kernel-based metric learning
  methods,''
\newblock in {\em ECCV}. Springer, 2014, pp. 1--16.

\bibitem{liao2015efficient}
Shengcai Liao and Stan~Z Li,
\newblock ``Efficient psd constrained asymmetric metric learning for person
  re-identification,''
\newblock in {\em ICCV}, 2015, pp. 3685--3693.

\bibitem{chen2016similarity}
Dapeng Chen, Zejian Yuan, Badong Chen, and Nanning Zheng,
\newblock ``Similarity learning with spatial constraints for person
  re-identification,''
\newblock in {\em CVPR}, 2016, pp. 1268--1277.

\bibitem{li2013locally}
Wei Li and Xiaogang Wang,
\newblock ``Locally aligned feature transforms across views,''
\newblock in {\em CVPR}, 2013, pp. 3594--3601.

\bibitem{matsukawa2016hierarchical}
Tetsu Matsukawa, Takahiro Okabe, Einoshin Suzuki, and Yoichi Sato,
\newblock ``Hierarchical gaussian descriptor for person re-identification,''
\newblock in {\em CVPR}, 2016, pp. 1363--1372.

\bibitem{he2016deep}
Kaiming He, Xiangyu Zhang, Shaoqing Ren, and Jian Sun,
\newblock ``Deep residual learning for image recognition,''
\newblock in {\em CVPR}, 2016, pp. 770--778.

\bibitem{li2014deepreid}
Wei Li, Rui Zhao, Tong Xiao, and Xiaogang Wang,
\newblock ``Deepreid: Deep filter pairing neural network for person
  re-identification,''
\newblock in {\em CVPR}, 2014, pp. 152--159.

\bibitem{ahmed2015improved}
Ejaz Ahmed, Michael Jones, and Tim~K Marks,
\newblock ``An improved deep learning architecture for person
  re-identification,''
\newblock in {\em CVPR}, 2015, pp. 3908--3916.

\bibitem{varior2016gated}
Rahul~Rama Varior, Mrinal Haloi, and Gang Wang,
\newblock ``Gated siamese convolutional neural network architecture for human
  re-identification,''
\newblock in {\em ECCV}. Springer, 2016, pp. 791--808.

\bibitem{lin2017consistent}
Ji~Lin, Liangliang Ren, Jiwen Lu, Jianjiang Feng, and Jie Zhou,
\newblock ``Consistent-aware deep learning for person re-identification in a
  camera network,''
\newblock in {\em CVPR}, 2017, pp. 5771--5780.

\bibitem{wang2014person}
Taiqing Wang, Shaogang Gong, Xiatian Zhu, and Shengjin Wang,
\newblock ``Person re-identification by video ranking,''
\newblock in {\em ECCV}. Springer, 2014, pp. 688--703.

\bibitem{tao2013person}
Dapeng Tao, Lianwen Jin, Yongfei Wang, Yuan Yuan, and Xuelong Li,
\newblock ``Person re-identification by regularized smoothing kiss metric
  learning,''
\newblock {\em IEEE TCSVT}, vol. 23, no. 10, pp. 1675--1685, 2013.

\bibitem{mclaughlin2016recurrent}
Niall McLaughlin, Jesus~Martinez Del~Rincon, and Paul Miller,
\newblock ``Recurrent convolutional network for video-based person
  re-identification,''
\newblock in {\em CVPR}, 2016, pp. 1325--1334.

\bibitem{zhou2017see}
Zhen Zhou, Yan Huang, Wei Wang, Liang Wang, and Tieniu Tan,
\newblock ``See the forest for the trees: Joint spatial and temporal recurrent
  neural networks for video-based person re-identification,''
\newblock in {\em CVPR}, 2017, pp. 4747--4756.

\bibitem{xu2017jointly}
Shuangjie Xu, Yu~Cheng, Kang Gu, Yang Yang, Shiyu Chang, and Pan Zhou,
\newblock ``Jointly attentive spatial-temporal pooling networks for video-based
  person re-identification,''
\newblock in {\em ICCV}, 2017, pp. 4733--4742.

\bibitem{liu2019spatial}
Yiheng Liu, Zhenxun Yuan, Wengang Zhou, and Houqiang Li,
\newblock ``Spatial and temporal mutual promotion for video-based person
  re-identification,''
\newblock in {\em AAAI}, 2019, vol.~33, pp. 8786--8793.

\bibitem{deng2014pedestrian}
Yubin Deng, Ping Luo, Chen~Change Loy, and Xiaoou Tang,
\newblock ``Pedestrian attribute recognition at far distance,''
\newblock in {\em ACM MM}, 2014, pp. 789--792.

\bibitem{liu2017hydraplus}
Xihui Liu, Haiyu Zhao, Maoqing Tian, Lu~Sheng, Jing Shao, Shuai Yi, Junjie Yan,
  and Xiaogang Wang,
\newblock ``Hydraplus-net: Attentive deep features for pedestrian analysis,''
\newblock in {\em ICCV}, 2017, pp. 350--359.

\bibitem{zhao2018grouping}
Xin Zhao, Liufang Sang, Guiguang Ding, Yuchen Guo, and Xiaoming Jin,
\newblock ``Grouping attribute recognition for pedestrian with joint recurrent
  learning.,''
\newblock in {\em IJCAI}, 2018, pp. 3177--3183.

\bibitem{gao2019pedestrian}
Lian Gao, Di~Huang, Yuanfang Guo, and Yunhong Wang,
\newblock ``Pedestrian attribute recognition via hierarchical multi-task
  learning and relationship attention,''
\newblock in {\em ACM MM}, 2019, pp. 1340--1348.

\bibitem{layne2012person}
Ryan Layne, Timothy~M Hospedales, Shaogang Gong, and Q~Mary,
\newblock ``Person re-identification by attributes.,''
\newblock in {\em BMVC}, 2012, vol.~2, p.~8.

\bibitem{li2014clothing}
Annan Li, Luoqi Liu, Kang Wang, Si~Liu, and Shuicheng Yan,
\newblock ``Clothing attributes assisted person reidentification,''
\newblock {\em IEEE TCSVT}, vol. 25, no. 5, pp. 869--878, 2014.

\bibitem{su2017multi}
Chi Su, Fan Yang, Shiliang Zhang, Qi~Tian, Larry~Steven Davis, and Wen Gao,
\newblock ``Multi-task learning with low rank attribute embedding for
  multi-camera person re-identification,''
\newblock {\em IEEE TPAMI}, vol. 40, no. 5, pp. 1167--1181, 2017.

\bibitem{wang2018transferable}
Jingya Wang, Xiatian Zhu, Shaogang Gong, and Wei Li,
\newblock ``Transferable joint attribute-identity deep learning for
  unsupervised person re-identification,''
\newblock in {\em CVPR}, 2018, pp. 2275--2284.

\bibitem{lin2019improving}
Yutian Lin, Liang Zheng, Zhedong Zheng, Yu~Wu, Zhilan Hu, Chenggang Yan, and
  Yi~Yang,
\newblock ``Improving person re-identification by attribute and identity
  learning,''
\newblock {\em Pattern Recognition}, vol. 95, pp. 151--161, 2019.

\bibitem{ling2019improving}
Hefei Ling, Ziyang Wang, Ping Li, Yuxuan Shi, Jiazhong Chen, and Fuhao Zou,
\newblock ``Improving person re-identification by multi-task learning,''
\newblock {\em Neurocomputing}, vol. 347, pp. 109--118, 2019.

\bibitem{han2018attribute}
Kai Han, Jianyuan Guo, Chao Zhang, and Mingjian Zhu,
\newblock ``Attribute-aware attention model for fine-grained representation
  learning,''
\newblock in {\em ACM MM}, 2018, pp. 2040--2048.

\bibitem{song2019two}
Wanru Song, Jieying Zheng, Yahong Wu, Changhong Chen, and Feng Liu,
\newblock ``A two-stage attribute-constraint network for video-based person
  re-identification,''
\newblock {\em IEEE Access}, vol. 7, pp. 8508--8518, 2019.

\bibitem{zhao2019attribute}
Yiru Zhao, Xu~Shen, Zhongming Jin, Hongtao Lu, and Xian-sheng Hua,
\newblock ``Attribute-driven feature disentangling and temporal aggregation for
  video person re-identification,''
\newblock in {\em CVPR}, 2019, pp. 4913--4922.

\bibitem{li2020appearance}
Shuzhao Li, Huimin Yu, and Haoji Hu,
\newblock ``Appearance and motion enhancement for video-based person
  re-identification,''
\newblock in {\em AAAI}, 2020, vol.~34, pp. 11394--11401.

\bibitem{yang2020spatial}
Jinrui Yang, Wei-Shi Zheng, Qize Yang, Ying-Cong Chen, and Qi~Tian,
\newblock ``Spatial-temporal graph convolutional network for video-based person
  re-identification,''
\newblock in {\em CVPR}, 2020, pp. 3289--3299.

\bibitem{hou2020temporal}
Ruibing Hou, Hong Chang, Bingpeng Ma, Shiguang Shan, and Xilin Chen,
\newblock ``Temporal complementary learning for video person
  re-identification,''
\newblock in {\em ECCV}. Springer, 2020, pp. 388--405.

\bibitem{chen2020temporal}
Guangyi Chen, Yongming Rao, Jiwen Lu, and Jie Zhou,
\newblock ``Temporal coherence or temporal motion: Which is more critical for
  video-based person re-identification?,''
\newblock in {\em ECCV}. Springer, 2020, pp. 660--676.

\bibitem{gu2020appearance}
Xinqian Gu, Hong Chang, Bingpeng Ma, Hongkai Zhang, and Xilin Chen,
\newblock ``Appearance-preserving 3d convolution for video-based person
  re-identification,''
\newblock in {\em ECCV}. Springer, 2020, pp. 228--243.

\bibitem{yan2020learning}
Yichao Yan, Jie Qin, Jiaxin Chen, Li~Liu, Fan Zhu, Ying Tai, and Ling Shao,
\newblock ``Learning multi-granular hypergraphs for video-based person
  re-identification,''
\newblock in {\em CVPR}, 2020, pp. 2899--2908.

\bibitem{zheng2016mars}
Liang Zheng, Zhi Bie, Yifan Sun, Jingdong Wang, Chi Su, Shengjin Wang, and
  Qi~Tian,
\newblock ``Mars: A video benchmark for large-scale person re-identification,''
\newblock in {\em ECCV}. Springer, 2016, pp. 868--884.

\bibitem{ristani2016performance}
Ergys Ristani, Francesco Solera, Roger Zou, Rita Cucchiara, and Carlo Tomasi,
\newblock ``Performance measures and a data set for multi-target, multi-camera
  tracking,''
\newblock in {\em ECCV}. Springer, 2016, pp. 17--35.

\bibitem{zhang2020multi}
Zhizheng Zhang, Cuiling Lan, Wenjun Zeng, and Zhibo Chen,
\newblock ``Multi-granularity reference-aided attentive feature aggregation for
  video-based person re-identification,''
\newblock in {\em CVPR}, 2020, pp. 10407--10416.

\bibitem{subramaniam2019co}
Arulkumar Subramaniam, Athira Nambiar, and Anurag Mittal,
\newblock ``Co-segmentation inspired attention networks for video-based person
  re-identification,''
\newblock in {\em ICCV}, 2019, pp. 562--572.

\bibitem{chen2020deep}
Jiaxin Chen, Jie Qin, Yichao Yan, Lei Huang, Li~Liu, Fan Zhu, and Ling Shao,
\newblock ``Deep local binary coding for person re-identification by delving
  into the details,''
\newblock in {\em ACM MM}, 2020, pp. 3034--3043.

\bibitem{vaswani2017attention}
Ashish Vaswani, Noam Shazeer, Niki Parmar, Jakob Uszkoreit, Llion Jones,
  Aidan~N Gomez, Lukasz Kaiser, and Illia Polosukhin,
\newblock ``Attention is all you need,''
\newblock {\em arXiv preprint arXiv:1706.03762}, 2017.

\bibitem{chen2019temporal}
Zhiyuan Chen, Annan Li, and Yunhong Wang,
\newblock ``A temporal attentive approach for video-based pedestrian attribute
  recognition,''
\newblock in {\em PRCV}. Springer, 2019, pp. 209--220.

\bibitem{chen2018deep}
Pu~Chen, Xinyi Xu, and Cheng Deng,
\newblock ``Deep view-aware metric learning for person re-identification.,''
\newblock in {\em IJCAI}, 2018, pp. 620--626.

\bibitem{zhu2020aware}
Zhihui Zhu, Xinyang Jiang, Feng Zheng, Xiaowei Guo, Feiyue Huang, Xing Sun, and
  Weishi Zheng,
\newblock ``Aware loss with angular regularization for person
  re-identification,''
\newblock in {\em AAAI}, 2020, vol.~34, pp. 13114--13121.

\bibitem{ye2021deep}
Mang Ye, Jianbing Shen, Gaojie Lin, Tao Xiang, Ling Shao, and Steven~CH Hoi,
\newblock ``Deep learning for person re-identification: A survey and outlook,''
\newblock {\em IEEE TPAMI}, 2021.

\bibitem{wen2016discriminative}
Yandong Wen, Kaipeng Zhang, Zhifeng Li, and Yu~Qiao,
\newblock ``A discriminative feature learning approach for deep face
  recognition,''
\newblock in {\em ECCV}. Springer, 2016, pp. 499--515.

\bibitem{luo2019bag}
Hao Luo, Youzhi Gu, Xingyu Liao, Shenqi Lai, and Wei Jiang,
\newblock ``Bag of tricks and a strong baseline for deep person
  re-identification,''
\newblock in {\em CVPR}, 2019, pp. 0--0.

\bibitem{zhang2020person}
Jianfu Zhang, Li~Niu, and Liqing Zhang,
\newblock ``Person re-identification with reinforced attribute attention
  selection,''
\newblock {\em IEEE TIP}, vol. 30, pp. 603--616, 2020.

\bibitem{deng2009imagenet}
Jia Deng, Wei Dong, Richard Socher, Li-Jia Li, Kai Li, and Li~Fei-Fei,
\newblock ``Imagenet: A large-scale hierarchical image database,''
\newblock in {\em CVPR}. Ieee, 2009, pp. 248--255.

\bibitem{li2018diversity}
Shuang Li, Slawomir Bak, Peter Carr, and Xiaogang Wang,
\newblock ``Diversity regularized spatiotemporal attention for video-based
  person re-identification,''
\newblock in {\em CVPR}, 2018, pp. 369--378.

\bibitem{bottou2010large}
L{\'e}on Bottou,
\newblock ``Large-scale machine learning with stochastic gradient descent,''
\newblock in {\em Proceedings of COMPSTAT'2010}, pp. 177--186. Springer, 2010.

\bibitem{kingma2014adam}
Diederik~P Kingma and Jimmy Ba,
\newblock ``Adam: A method for stochastic optimization,''
\newblock {\em arXiv preprint arXiv:1412.6980}, 2014.

\bibitem{sun2018beyond}
Yifan Sun, Liang Zheng, Yi~Yang, Qi~Tian, and Shengjin Wang,
\newblock ``Beyond part models: Person retrieval with refined part pooling (and
  a strong convolutional baseline),''
\newblock in {\em ECCV}, 2018, pp. 480--496.

\bibitem{fu2019sta}
Yang Fu, Xiaoyang Wang, Yunchao Wei, and Thomas Huang,
\newblock ``Sta: Spatial-temporal attention for large-scale video-based person
  re-identification,''
\newblock in {\em AAAI}, 2019, vol.~33, pp. 8287--8294.

\bibitem{hou2019vrstc}
Ruibing Hou, Bingpeng Ma, Hong Chang, Xinqian Gu, Shiguang Shan, and Xilin
  Chen,
\newblock ``Vrstc: Occlusion-free video person re-identification,''
\newblock in {\em CVPR}, 2019, pp. 7183--7192.

\bibitem{li2019global}
Jianing Li, Jingdong Wang, Qi~Tian, Wen Gao, and Shiliang Zhang,
\newblock ``Global-local temporal representations for video person
  re-identification,''
\newblock in {\em ICCV}, 2019, pp. 3958--3967.

\bibitem{wu2020adaptive}
Yiming Wu, Omar El~Farouk Bourahla, Xi~Li, Fei Wu, Qi~Tian, and Xue Zhou,
\newblock ``Adaptive graph representation learning for video person
  re-identification,''
\newblock {\em IEEE TIP}, vol. 29, pp. 8821--8830, 2020.

\bibitem{wu2018exploit}
Yu~Wu, Yutian Lin, Xuanyi Dong, Yan Yan, Wanli Ouyang, and Yi~Yang,
\newblock ``Exploit the unknown gradually: One-shot video-based person
  re-identification by stepwise learning,''
\newblock in {\em CVPR}, 2018, pp. 5177--5186.

\bibitem{liu2019spatially}
Chih-Ting Liu, Chih-Wei Wu, Yu-Chiang~Frank Wang, and Shao-Yi Chien,
\newblock ``Spatially and temporally efficient non-local attention network for
  video-based person re-identification,''
\newblock {\em arXiv preprint arXiv:1908.01683}, 2019.

\bibitem{liu2020temporal}
Jiawei Liu, Xierong Zhu, and Zheng-Jun Zha,
\newblock ``Temporal attribute-appearance learning network for video-based
  person re-identification,''
\newblock {\em arXiv preprint arXiv:2009.04181}, 2020.

\bibitem{van2008visualizing}
Laurens Van~der Maaten and Geoffrey Hinton,
\newblock ``Visualizing data using t-sne.,''
\newblock {\em JMLR}, vol. 9, no. 11, 2008.

\bibitem{zheng2019pose}
Liang Zheng, Yujia Huang, Huchuan Lu, and Yi~Yang,
\newblock ``Pose-invariant embedding for deep person re-identification,''
\newblock {\em IEEE TIP}, vol. 28, no. 9, pp. 4500--4509, 2019.

\bibitem{zheng2017person}
Liang Zheng, Hengheng Zhang, Shaoyan Sun, Manmohan Chandraker, Yi~Yang, and
  Qi~Tian,
\newblock ``Person re-identification in the wild,''
\newblock in {\em CVPR}, 2017, pp. 1367--1376.

\bibitem{hermans2017defense}
Alexander Hermans, Lucas Beyer, and Bastian Leibe,
\newblock ``In defense of the triplet loss for person re-identification,''
\newblock {\em arXiv preprint arXiv:1703.07737}, 2017.

\bibitem{chen2017beyond}
Weihua Chen, Xiaotang Chen, Jianguo Zhang, and Kaiqi Huang,
\newblock ``Beyond triplet loss: a deep quadruplet network for person
  re-identification,''
\newblock in {\em CVPR}, 2017, pp. 403--412.

\bibitem{chai2021semantically}
Tianrui Chai, Xinyu Mei, Annan Li, and Yunhong Wang,
\newblock ``Semantically-guided disentangled representation for robust gait
  recognition,''
\newblock in {\em ICME}. IEEE, 2021, pp. 1--6.

\bibitem{wang2021robust}
Zhikang Wang, Lihuo He, Xiaoguang Tu, Jian Zhao, Xinbo Gao, Shengmei Shen, and
  Jiashi Feng,
\newblock ``Robust video-based person re-identification by hierarchical
  mining,''
\newblock {\em IEEE TCSVT}, 2021.

\bibitem{li2020hierarchical}
Peike Li, Pingbo Pan, Ping Liu, Mingliang Xu, and Yi~Yang,
\newblock ``Hierarchical temporal modeling with mutual distance matching for
  video based person re-identification,''
\newblock {\em IEEE TCSVT}, vol. 31, no. 2, pp. 503--511, 2020.

\bibitem{zhang2018learning}
Wei Zhang, Shengnan Hu, Kan Liu, and Zhengjun Zha,
\newblock ``Learning compact appearance representation for video-based person
  re-identification,''
\newblock {\em IEEE TCSVT}, vol. 29, no. 8, pp. 2442--2452, 2018.

\bibitem{liu2018hierarchical}
Zheng Liu, Yunhong Wang, and Annan Li,
\newblock ``Hierarchical integration of rich features for video-based person
  re-identification,''
\newblock {\em IEEE TCSVT}, vol. 29, no. 12, pp. 3646--3659, 2018.

\bibitem{shi2020person}
Yuxuan Shi, Zhen Wei, Hefei Ling, Ziyang Wang, Jialie Shen, and Ping Li,
\newblock ``Person retrieval in surveillance videos via deep attribute mining
  and reasoning,''
\newblock {\em IEEE TMM}, 2020.

\bibitem{tay2019aanet}
Chiat-Pin Tay, Sharmili Roy, and Kim-Hui Yap,
\newblock ``Aanet: Attribute attention network for person re-identifications,''
\newblock in {\em CVPR}, 2019, pp. 7134--7143.

\bibitem{li2019attribute}
Huafeng Li, Shuanglin Yan, Zhengtao Yu, and Dapeng Tao,
\newblock ``Attribute-identity embedding and self-supervised learning for
  scalable person re-identification,''
\newblock {\em IEEE TCSVT}, vol. 30, no. 10, pp. 3472--3485, 2019.

\bibitem{wang2019learning}
Zheng Wang, Junjun Jiang, Yang Wu, Mang Ye, Xiang Bai, and Shin’ichi Satoh,
\newblock ``Learning sparse and identity-preserved hidden attributes for person
  re-identification,''
\newblock {\em IEEE TIP}, vol. 29, pp. 2013--2025, 2019.

\bibitem{li2020structure}
Huafeng Li, Zhenyu Kuang, Zhengtao Yu, and Jiebo Luo,
\newblock ``Structure alignment of attributes and visual features for
  cross-dataset person re-identification,''
\newblock {\em Pattern Recognition}, vol. 106, pp. 107414, 2020.

\bibitem{yan2021beyond}
Cheng Yan, Guansong Pang, Xiao Bai, Changhong Liu, Ning Xin, Lin Gu, and Jun
  Zhou,
\newblock ``Beyond triplet loss: person re-identification with fine-grained
  difference-aware pairwise loss,''
\newblock {\em IEEE TMM}, 2021.

\bibitem{zhu2020asta}
Xierong Zhu, Jiawei Liu, Haoze Wu, Meng Wang, and Zheng-Jun Zha,
\newblock ``Asta-net: Adaptive spatio-temporal attention network for person
  re-identification in videos,''
\newblock in {\em ACM MM}, 2020, pp. 1706--1715.

\bibitem{hou2021bicnet}
Ruibing Hou, Hong Chang, Bingpeng Ma, Rui Huang, and Shiguang Shan,
\newblock ``Bicnet-tks: Learning efficient spatial-temporal representation for
  video person re-identification,''
\newblock in {\em CVPR}, 2021, pp. 2014--2023.

\bibitem{liu2019person}
Zimo Liu, Huchuan Lu, Xiang Ruan, and Ming-Hsuan Yang,
\newblock ``Person reidentification by joint local distance metric and feature
  transformation,''
\newblock {\em IEEE TNNLS}, vol. 30, no. 10, pp. 2999--3009, 2019.

\bibitem{Zhao_2021_CVPR}
Jianan Zhao, Fengliang Qi, Guangyu Ren, and Lin Xu,
\newblock ``Phd learning: Learning with pompeiu-hausdorff distances for
  video-based vehicle re-identification,''
\newblock in {\em CVPR}, June 2021, pp. 2225--2235.

\bibitem{gong2021lag}
Xun Gong, Zu~Yao, Xin Li, Yueqiao Fan, Bin Luo, Jianfeng Fan, and Boji Lao,
\newblock ``Lag-net: Multi-granularity network for person re-identification via
  local attention system,''
\newblock {\em IEEE TMM}, 2021.

\bibitem{wu2018and}
Lin Wu, Yang Wang, Junbin Gao, and Xue Li,
\newblock ``Where-and-when to look: Deep siamese attention networks for
  video-based person re-identification,''
\newblock {\em IEEE TMM}, vol. 21, no. 6, pp. 1412--1424, 2018.

\bibitem{choi2021arm}
Yapkan Choi, Yeshwanth Napolean, and Jan~C van Gemert,
\newblock ``The arm-swing is discriminative in video gait recognition for
  athlete re-identification,''
\newblock {\em arXiv preprint arXiv:2106.11280}, 2021.

\end{thebibliography}

%








\end{document}